\documentclass[a4paper,fleqn]{cas-dc}
\usepackage[authoryear]{natbib}
\usepackage{subcaption}
\usepackage{multirow}
\usepackage{adjustbox}
\usepackage{amsmath,amsfonts,bm}
\usepackage{alphalph}
\usepackage{amssymb}% http://ctan.org/pkg/amssymb
\usepackage{pifont}% http://ctan.org/pkg/pifont
\def\tsc#1{\csdef{#1}{\textsc{\lowercase{#1}}\xspace}}
\tsc{WGM}
\tsc{QE}
\tsc{EP}
\tsc{PMS}
\tsc{BEC}
\tsc{DE}
\def\eqref#1{equation~\ref{#1}}
\def\1{\bm{1}}

\newcommand{\model}{}

\begin{document}
\let\WriteBookmarks\relax
\def\floatpagepagefraction{1}
\def\textpagefraction{.001}

\shorttitle{\model}

\shortauthors{Xia et~al.}

\title [mode = title]{Cross-View Geo-Localization with Street-View and VHR Satellite Imagery in Decentrality Settings}                      
% \tnotemark[1,2]

% \tnotetext[1]{This document is the results of the research
%   project funded by the National Science Foundation.}

% \tnotetext[2]{The second title footnote which is a longer text matter
%   to fill through the whole text width and overflow into
%   another line in the footnotes area of the first page.}

\author[1]{Panwang Xia}[type=editor,
                        auid=000,
                        bioid=1]

\author[2]{Lei Yu}[type=editor,
                        auid=000,
                        bioid=1,
                        % prefix=Sir,
                        % role=Researcher,
                        ]

\author[1,3]{Yi Wan}[type=editor,
                        auid=000,
                        bioid=1,
                        % prefix=Sir,
                        % role=Researcher,
                        orcid=0000-0001-6777-6047
                        ]
\cormark[1]
\ead{yi.wan@whu.edu.cn}

\author[1]{Qiong Wu}[type=editor,
                        auid=000,
                        bioid=1,
                        % prefix=Sir,
                        % role=Researcher,
                        ]

\author[1]{Peiqi Chen}[type=editor,
                        auid=000,
                        bioid=1,
                        % prefix=Sir,
                        % role=Researcher,
                        ]

\author[2]{Liheng Zhong}[type=editor,
                        auid=000,
                        bioid=1,
                        % prefix=Sir,
                        % role=Researcher,
                        ]

\author[1]{Yongxiang Yao}[type=editor,
                        auid=000,
                        bioid=1,
                        % prefix=Sir,
                        % role=Researcher,
                        ]

\author[1]{Dong Wei}[type=editor,
                        auid=000,
                        bioid=1,
                        % prefix=Sir,
                        % role=Researcher,
                        ]

\author[1,3]{Xinyi Liu}[type=editor,
                        auid=000,
                        bioid=1,
                        % prefix=Sir,
                        % role=Researcher,
                        ]

\author[2]{Lixiang Ru}[type=editor,
                        auid=000,
                        bioid=1,
                        % prefix=Sir,
                        % role=Researcher,
                        ]

\author[2]{Yingying Zhang}[type=editor,
                        auid=000,
                        bioid=1,
                        % prefix=Sir,
                        % role=Researcher,
                        ]

\author[2]{Jiangwei Lao}[type=editor,
                        auid=000,
                        bioid=1,
                        % prefix=Sir,
                        % role=Researcher,
                        ]

\author[2]{Jingdong Chen}[type=editor,
                        auid=000,
                        bioid=1,
                        % prefix=Sir,
                        % role=Researcher,
                        ]

\author[2]{Ming Yang}[type=editor,
                        auid=000,
                        bioid=1,
                        % prefix=Sir,
                        % role=Researcher,
                        ]

\author[1,3]{Yongjun Zhang}[type=editor,
                        auid=000,
                        bioid=1,
                        % prefix=Sir,
                        % role=Researcher,
                        orcid=0000-0001-9845-4251
                        ]
\cormark[1]
\ead{zhangyj@whu.edu.cn}

\affiliation[1]{organization={School of Remote Sensing and Information Engineering, Wuhan University},%Department and Organization
            city={Wuhan},
            postcode={430079}, 
            state={Hubei},
            country={China}}

\affiliation[2]{organization={Ant Group},%Department and Organization
            city={Hangzhou},
            postcode={310023}, 
            state={Zhejiang},
            country={China}}

\affiliation[3]{organization={Technology Innovation Center for Collaborative Applications of Natural Resources Data in GBA, Ministry of Natural Resources},%Department and Organization
            city={Guangzhou},
            postcode={510075}, 
            state={Guangdong},
            country={China}}

\cortext[cor1]{Corresponding authors}
% \cortext[cor2]{Principal corresponding author}

% \fntext[fn1]{Both authors contribute equally to this work.}
% \fntext[fn2]{Work done while working at mosaix.ai}
% \fntext[fn2]{Another author footnote, this is a very long footnote and
%   it should be a really long footnote. But this footnote is not yet
%   sufficiently long enough to make two lines of footnote text.}

% % For a title note without a number/mark
% \nonumnote{This note has no numbers. In this work we demonstrate $a_b$
%   the formation Y\_1 of a new type of polariton on the interface
%   between a cuprous oxide slab and a polystyrene micro-sphere placed

\begin{abstract}
Cross-View Geo-Localization tackles the challenge of image geo-localization in GNSS-denied environments, including disaster response scenarios, urban canyons, and dense forests, by matching street-view query images with geo-tagged aerial-view reference images. However, current research often relies on benchmarks and methods that assume center-aligned settings or account for only limited \textbf{decentrality}, which we define as the offset of the query image relative to the reference image center. Such assumptions fail to reflect real-world scenarios, where reference databases are typically pre-established without the possibility of ensuring perfect alignment for each query image. Moreover, decentrality is a critical factor warranting deeper investigation, as larger decentrality can substantially improve localization efficiency but comes at the cost of declines in localization accuracy. To address this limitation, we introduce DReSS (\textbf{D}ecentrality \textbf{Re}lated \textbf{S}treet-view and \textbf{S}atellite-view dataset), a novel dataset designed to evaluate cross-view geo-localization with a large geographic scope and diverse landscapes, emphasizing the decentrality issue. Meanwhile, we propose AuxGeo (\textbf{Aux}iliary Enhanced \textbf{Geo}-Localization) to further study the decentrality issue, which leverages a multi-metric optimization strategy with two novel modules: the Bird's-eye view Intermediary Module (BIM) and the Position Constraint Module (PCM). BIM uses bird's-eye view images derived from street-view panoramas as an intermediary, simplifying the “cross-view challenge with decentrality” to a cross-view problem and a decentrality problem. PCM leverages position priors between cross-view images to establish multi-grained alignment constraints. These modules improve the localization accuracy despite the decentrality problem. Extensive experiments demonstrate that AuxGeo outperforms previous methods on our proposed DReSS dataset, mitigating the issue of large decentrality, and also achieves state-of-the-art performance on existing public datasets such as CVUSA, CVACT, and VIGOR. The codes and dataset will be made available at \href{https://github.com/SummerpanKing/DReSS}{https://github.com/SummerpanKing/DReSS}.

\end{abstract}

% \includegraphics{figs/grabs.pdf}

%\input{highlight.tex}

% Keywords

\maketitle

\section{Introduction}

The cross-view coupling of street-view and satellite-view imagery in the spatial dimension is a prominent research focus in the field of remote sensing. This coupling not only facilitates better perspective observation of Earth's surface \citep{li2023omnicity, ye2024sg} but also assists navigation in scenarios where GNSS signals are unavailable. Cross-View Geo-Localization (CVGL) addresses the challenge of image geo-localization by matching street-view query images with geo-tagged very high-resolution (VHR) satellite reference images from pre-established databases. It is particularly beneficial in complex environments where GNSS signals are unavailable, such as urban canyons with dense building clusters or heavily forested areas \cite{ye2024coarse}. This technology offers us an alternative way to localize ourselves in real scenarios, enabling emerging applications such as autonomous driving \citep{hane20173d,kim2017satellite,wan2016illumination}, robotic navigation \citep{mcmanus2014shady}, augmented reality \citep{chiu2018augmented}, geographic information aggregation \citep{he2024advlut}, and disaster response \citep{li2024cross}.

\begin{figure}
    \centering
    \includegraphics[width=1\linewidth]{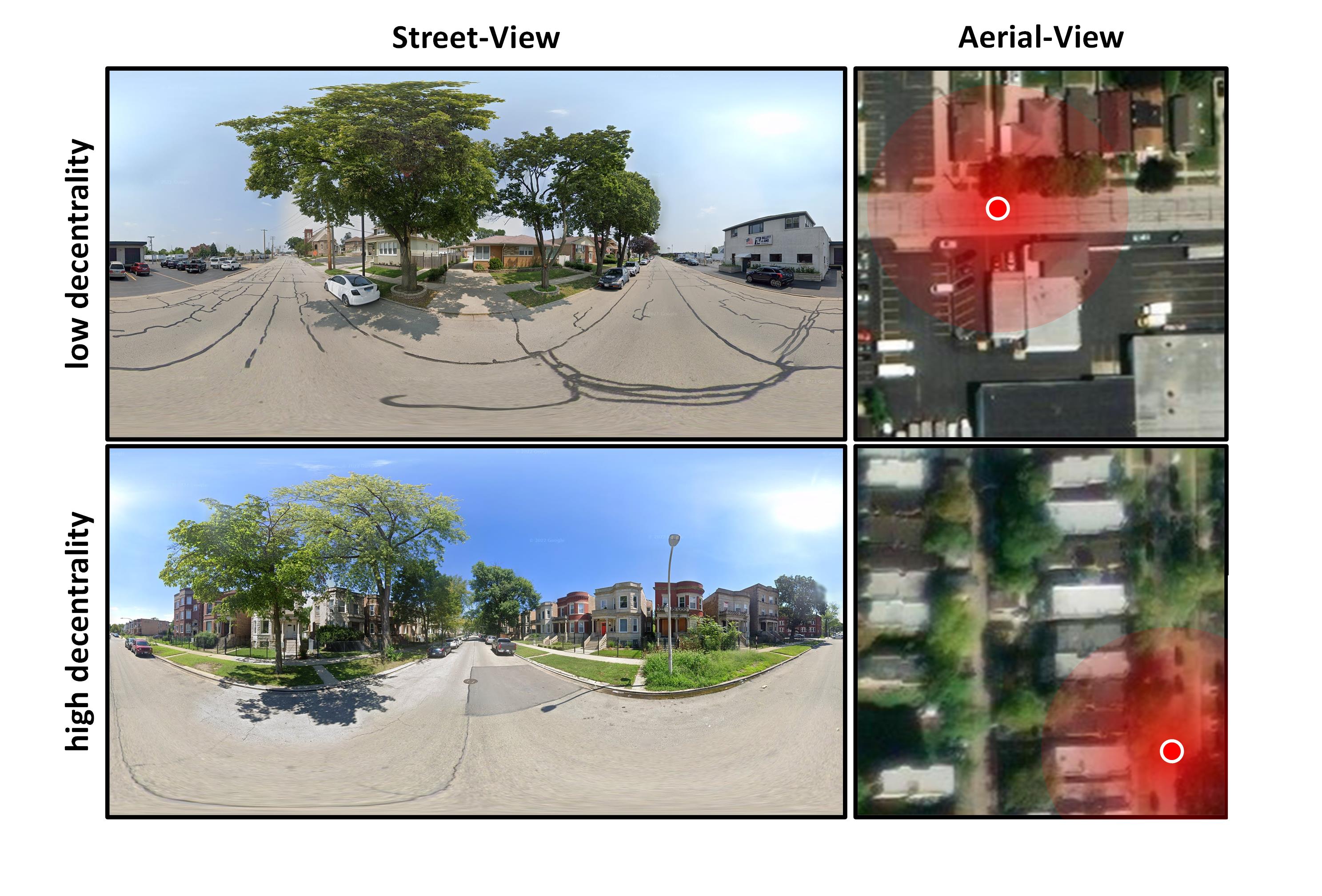}
    \caption{Visualization of the decentrality issue. Red circles simulate the visible regions of street-view panoramas in VHR satellite reference images. Higher decentrality reduces global similarity, increasing the difficulty of establishing cross-view image correspondence.}
    \label{fig:decentrality_vis_perceptron}
\end{figure}

CVGL is challenging primarily due to the significant appearance changes between street-view and aerial-view images caused by viewpoint variations \citep{ling2022graph,elhashash2022cross}. Moreover, in real-world applications, it is not likely to have perfectly matched (center-aligned) reference images available for each query image, since the reference database is established with features generated by aerial images in advance. CVGL has to be robust and able to tolerate offsets between a street-view query and its corresponding aerial-view reference. Thus, we introduce the concept of \emph{decentrality}, measuring the offset level between a query image position w.r.t. the center of its reference image. For the same search region, large decentrality requires less overlap between reference images, significantly reducing the number of reference images in the database and thereby improving the retrieval efficiency, which is essential for localization tasks. However, large decentrality presents substantial challenges to CVGL by diminishing content similarity between cross-view images and introducing disruptions caused by non-co-visible elements. Addressing these challenges requires robust partial matching mechanisms to effectively align cross-view images, as demonstrated in Figure \ref{fig:decentrality_vis_perceptron}. A detailed statistical analysis included in section \ref{decentrality_discription} also highlights the importance of the decentrality issue in real-world applications.

In existing datasets like CVUSA \citep{zhai2017predicting} and CVACT \citep{liu2019lending}, the correspondences between cross-view images are assumed to be well center-aligned, which is over-simplified for real-world applications. The VIGOR dataset \citep{zhu2021vigor} relaxes this restriction of center-aligned correspondence to some extent, while, a query image is still located within a quite small area around the center of its reference image, see Figure \ref{fig:dataset_setting}(a). While the VIGOR dataset defines some "semi-positive" street-view images with larger decentrality, there is always an alternative reference image available to form a "positive" pair for every street-view image. As a result, the VIGOR dataset is not well-suited for effectively investigating the decentrality issue, even if its original structure is reorganized. Therefore, we propose a novel dataset, DReSS, to evaluate CVGL in more realistic settings. DReSS distinguishes itself from previous datasets in two key ways. First, it allows for the evaluation of CVGL with larger decentrality (see Figure \ref{fig:dataset_setting}(a)) by sampling reference images seamlessly with a low overlap (12.5\%). Second, DReSS comprises cross-view images collected from 8 cities worldwide, covering both urban and suburban areas with diverse landscapes, where the area of each reference image exceeds 400 \(km^2\). This new large dataset enables a step forward in studying CVGL in a practical setting. A detailed comparison of decentrality among the above datasets is included in section \ref{comparison_datasets}.

Large decentrality presents significant challenges for existing methods. Methods using polar transform, which convert aerial-view images into street-view-like images \citep{shi2019spatial}, are unsuitable without center-aligned settings. Other state-of-the-art methods \citep{deuser2023sample4geo,zhang2024aligning} do not take into consideration the impact of offsets between street-view and aerial-view images across a broader portion of decentrality, resulting in a significant decrease in retrieval precision. Therefore, we propose a novel method, AuxGeo (auxiliary enhanced geolocalization), to address the aforementioned problems. The AuxGeo employs multi-metric optimization with two innovative modules incorporated as auxiliary tasks to enhance the representation ability of the backbone. To mitigate viewpoint variation involving decentrality, we introduce the Bird's-eye view Intermediary Module (BIM). The BIM uses BEV images derived from street-view panoramas as intermediaries, establishing additional connections between street-BEV and aerial-BEV images, simplifying the “cross-view challenge with decentrality” to a cross-view problem and a decentrality problem. To mitigate the problem of decentrality, we introduce the Position Constraint Module (PCM). The PCM leverages prior position correspondences between street-view and aerial-view images as supervision, establishing alignments in a multi-grained manner and enhancing the backbone's ability to learn cross-view correspondences for feature representation even with significant decentrality. 

Extensive experiments demonstrate that AuxGeo improves retrieval precision, surpassing previous methods on our proposed DReSS dataset and existing public datasets. Our main contributions can be summarized as follows:
\begin{itemize}
\item We introduce the decentrality issue in CVGL, which emphasizes the practical challenges caused by positional offsets between street-view queries and their corresponding aerial-view references. 
\item We propose DReSS, the first public dataset designed to comprehensively evaluate CVGL methods in decentrality settings. It includes cross-view images from diverse geographic locations with extensive land coverage.
\item We propose a multi-metric optimization-based method, AuxGeo, with two novel modules, BIM and PCM, that effectively address the cross-view problem and mitigate the decentrality problem for cross-view geo-localization.
\item Our proposed method AuxGeo outperforms previous approaches on the DReSS dataset, showing progressively greater improvements over the current methods as decentrality increases. It also achieves state-of-the-art localization accuracy on existing public datasets. Importantly, AuxGeo incurs no additional pre-processes or computational costs during inference.
\end{itemize}

The remainder of this paper is structured as follows: Section \ref{related work} provides a comprehensive review of related work, including existing datasets and geo-localization methods in the cross-view topic. Section \ref{DReSS dataset} introduces the proposed DReSS dataset in detail and offers an in-depth analysis of the importance the decentrality issue. Section \ref{Methodology} outlines the proposed method, AuxGeo. Experimental results and discussions are presented in Sections \ref{Experiments} and \ref{Discussion}, respectively. Finally, Section \ref{Conclusion} concludes the paper with an analysis of the method's advantages and limitations.
 
\section{Related Work}
\label{related work}

%% Use \subsection commands to start a subsection.
\subsection{Cross-View Datasets}
There are many datasets that have been introduced for cross-view geo-localization in the past decade \citep{lin2013cross,vo2016localizing,tian2017cross}. Among them, the most widely used datasets are CVUSA, CVACT, VIGOR and University-1652.

The original CVUSA is a huge dataset containing over 1 million ground and aerial images from multiple cities in the United States, first proposed by Workman et al. 
\citep{workman2015wide}. The current used version of CVUSA is a subset made by Zhai et al. \citep{zhai2017predicting}, including 35,532 street-aerial image pairs for training and 8,884 pairs for testing. Similar to CVUSA, Liu et al. \citep{liu2019lending} proposed CVACT which has 35,532 image pairs for training, 8,884 pairs for validation and 92,802 pairs for testing. Extending beyond conventional one-to-one retrieval, Zhu et al. \citep{zhu2021vigor} introduced VIGOR. This dataset comprises 105,214 street-view images and 90,618 aerial-view images, uniquely characterized by its assumption of random placements within the target area without center-aligned settings. VIGOR is regarded as the most realistic dataset for cross-view geo-localization, as it extends the relationships between cross-view images beyond merely center-aligned correspondence. However, even with this advancement, the decentrality of positive samples in VIGOR is still confined to a relatively centered area, which restricts its potential for a more thorough exploration of the issue. University-1652 \citep{zheng2020university} is the first dataset to simultaneously include images from satellite, synthetic drone, and ground levels, providing 1,652 cross-view image sets from universities around the world. In recent months, several datasets \citep{ye2025cross, huang2024cv} have been proposed. However, all of them fail to address the issue of decentrality. To overcome this limitation, we propose DReSS, which provides a wider range of decentrality, facilitating a more in-depth investigation.

\subsection{Cross-View Geo-Localization}
In early works, researchers trained a two-stream CNN to extract embedding representations from street-view and aerial-view images. The performance of these methods has improved compared with handcraft methods but still suffers from dramatic viewpoint variation. Current methods can be roughly categorized into geometry-based and feature-based methods.

\subsubsection{Geometry-based Cross-View Geo-Localization.} To bridge the cross-view discrepancy, Shi et al. \citep{shi2019spatial} first utilized polar transform on aerial-view images to achieve perspective transformation using geometry. This approach has since been widely adopted in subsequent research \citep{shi2020looking,yang2021cross,wang2023dehi,zhang2023cross}. Nevertheless, the initial application of the polar transform introduced significant distortions in the visual appearance. To counteract these distortions, Generative Adversarial Networks (GANs) have been employed, demonstrating efficacy in restoring the original appearance of transformed images \citep{toker2021coming,regmi2018cross,regmi2019bridging,lu2020geometry,shi2022geometry}. With the introduction of the VIGOR \citep{zhu2021vigor}, the polar transform-based methods are no longer applicable to non-center-aligned cross-view image pairs. Zhang et al. \citep{zhang2024aligning} proposed a feature recombination method that uses geometric spatial layout correspondence between cross-views to replace the polar transform. Though it has good performance on VIGOR, the design is not suitable for datasets with larger range of decentrality, like the proposed DReSS.

\subsubsection{Non-Geometry-based Cross-View Geo-Localization.} Simultaneously, methods aimed at bridging the cross-view gap directly through the enhancement of feature representation \citep{shi2019spatial,shi2020optimal,hu2018cvm,sun2019geocapsnet,ye2024coarse,cheng2018crowd,wu2024camp,xia2024enhancing} are also in widely research. Several works use ViT as a backbone \citep{yang2021cross,zhu2022transgeo,huang2024cv} for better feature extraction via encoding spatial information using self-attention mechanism. With the support of powerful backbone networks, recent feature-based methods \citep{deuser2023sample4geo} can achieve good performance, and some methods \citep{li2024unleashing,li2024learning} can even be trained in an unsupervised manner, albeit with limited performance. 

\section{ DReSS dataset}
\label{DReSS dataset}

\subsection{Problem Statement} 
\begin{figure}
    \centering
    \includegraphics[width=1\linewidth]{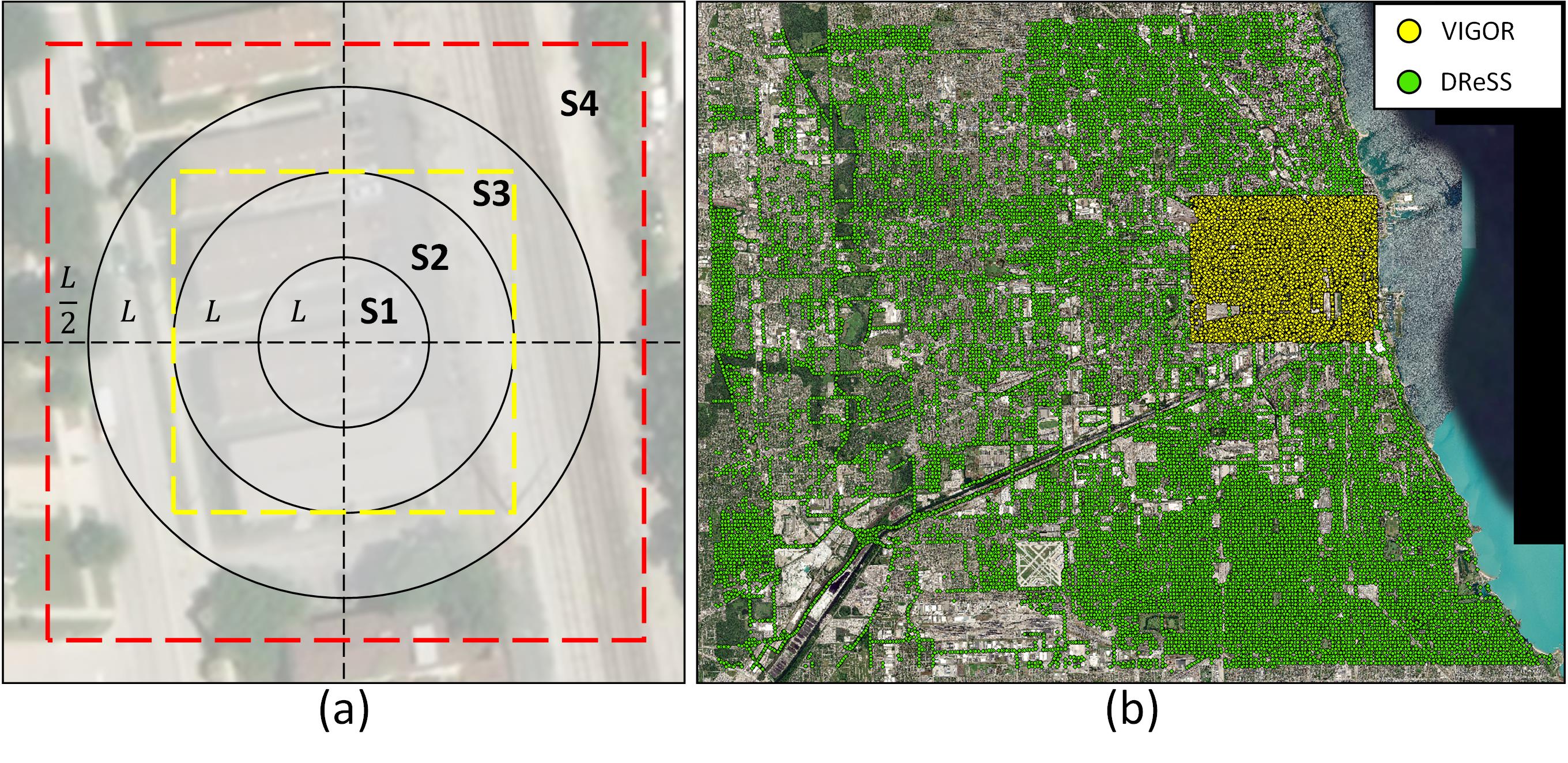}
    \caption{(a) Comparison of the hit area of VIGOR (yellow box) and DReSS (red box). Four subsets are divided within the hit area of DReSS with rising decentrality. (b) Comparison of the coverage scope between VIGOR and DReSS.}
    \label{fig:dataset_setting}
\end{figure}

Given an area of interest (AOI), the goal of cross-view geo-localization is to determine the location of a street-view image arbitrarily within it by establishing correspondence with geo-tagged aerial-view images. Beyond previous datasets, the DReSS goes further in investigating the decentrality between cross-view images. As depicted in Figure \ref{fig:dataset_setting}(a), we define the area of the best matched street-view panorama of an aerial-view reference image as the \emph{hit area}. Compared to VIGOR's hit area (depicted as the yellow box), the hit area in DReSS (depicted as the red box) is notably larger, achieved through a lower overlap ratio of 12.5\% (versus 50\% in VIGOR). This disparity indicates a wider portion of decentrality among cross-view images in DReSS. 

\begin{table}[h]
    \centering
 
    \begin{tabular}{c|cccc} 
    \toprule
    \toprule
         Subset&   S1&   S2& S3&S4\\ 
    \midrule
         Train&  5,704&  16,965& 28,059&36,737\\ 
         Test& 5,716& 17,146& 28,056&36,551\\
    \bottomrule
    \bottomrule
    \end{tabular}
    \caption{Number of best-matched cross-view image pairs in four subsets.}
    \label{tab:subsets distribution}
\end{table}

To thoroughly investigate the impact of decentrality on cross-view geo-localization, we subdivided the red box into four subsets, labeled S1 through S4. Each subset represents progressively increasing degrees of decentrality and with no interaction. We counted the number of best-matched cross-view image pairs in Subsets 1 to 4, for both the training and test settings. As shown in Table \ref{tab:subsets distribution}, the number of image pairs in S1 to S4 corresponds to their respective sizes, confirming the uniform distribution.

\subsection{Decentrality Issue}
\label{decentrality_discription}
In real-world applications of cross-view geo-localization, determining the overlap among reference images is crucial for constructing an effective reference image database, as it directly influences the database size, retrieval precision, and time consumption of the geo-localization process. However, this aspect has been understudied, and there is no unified industry standard. To address this gap, we conducted a comprehensive statistical analysis of various overlap levels among reference databases, assessing their impact on the best-matched cross-view image pairs across all geo-locations within the DReSS dataset. 

\begin{table*}[h]
    \centering
    \begin{tabular}{cc|cccc|cc} 
    \toprule
    \toprule
          Overlap&Subset&   S1&   S2& S3& S4& All& Ratio\\ 
    \midrule
          12.5\%&Train&  5,704&  16,965& 28,059&36,737 & 422,760&1x\\ 
          DReSS&Test& 5,716& 17,146& 28,056&36,551 & &\\
        \midrule
 20\%& Train& 6,797& 20,114& 33,755&26,916 & 563,714&1.33x\\
 & Test& 6,725& 20,451& 33,376&27,036 & &\\
 \midrule
 30\%& Train& 9,195& 26,847& 40,065&11,475 & 731,871&1.73x\\
 & Test& 9,138& 26,713& 40,155&11,582 & &\\
 \midrule
 40\%& Train& 11,947& 36,017& 37,189&2,429 & 1,003,750&2.37x\\
 & Test& 12,080& 35,819& 37,311&2,378 & &\\
 \midrule
 50\%& Train& 17,240& 51,541& 18,798&3 & 1,436,317&3.40x\\
 & Test& 17,167& 51,720& 18,694&7 & &\\
 \bottomrule
 \bottomrule
    \end{tabular}
    \caption{Statistical analysis of the reference dataset under varying overlap levels. S1-S4 represent the number of best-matched cross-view image pairs across four subsets, corresponding to different levels of decentrality. "All" denotes the total number of images in the database, while "Ratio" indicates the relative size compared to the baseline overlap of 12.5\%. }
    \label{tab:subsets_distribution}
\end{table*}

We evaluated a range of overlap levels: 12.5\%, 20\%, 30\%, 40\%, and 50\% among the reference aerial-view images. Maintaining the subset configurations from DReSS, as illustrated in Figure \ref{fig:dataset_setting}(a), we recorded the number of cross-view image pairs within subsets 1 to 4, each representing increasing levels of decentrality under different overlap levels. The results, detailed in Table \ref{tab:subsets_distribution}, reveal that in 3 out of 5 cases of overlap settings, image pairs with large decentrality constitute a significant portion. For overlap levels of 12.5\%, 20\%, and 30\%, image pairs in subset 4 (large decentrality) account for 13\%, 31\%, and 42\%, respectively. This distribution pattern underscores that decentrality is a critical issue and needs to be fully investigated in CVGL.

Furthermore, we recorded the size of the reference database (number of reference images) under different overlap levels, as shown in Table \ref{tab:subsets_distribution}. The results indicate that while higher overlap reduces decentrality, it also increases the size of the reference database, which in turn decreases the efficiency of cross-view geo-localization. This trade-off further emphasizes the need for a thorough investigation into decentrality.

\subsection{Comparison with Previous Datasets} 
\label{comparison_datasets}

\begin{table*}[ht]
    \centering
    
    \begin{tabular}{l|c|c|c|c|c}
    \toprule
    \toprule
         &   CVUSA&CVACT&  VIGOR&CVGlobal&DReSS\\
        \midrule
         Reference images&   44,416&128,334&  90,618&134,233&422,760\\
         Query images&   44,416&128,334&  105,214&134,233&174,934\\
         Portion of decentrality&   None&None&  Narrow&None&Large\\
         Geo-location distribution&   USA&Austrilia&  USA&Worldwide&Worldwide\\
 Regions& Suburban& Urban& Urban&Urban&Urban and suburban\\
 \bottomrule
 \bottomrule
    \end{tabular}
    \caption{Comparison between the proposed DReSS dataset and existing datasets for cross-view geo-localization task.}
    \label{tab:head-to-head comparison}
\end{table*}

\begin{figure}[h]
    \centering
    \includegraphics[width=1\linewidth]{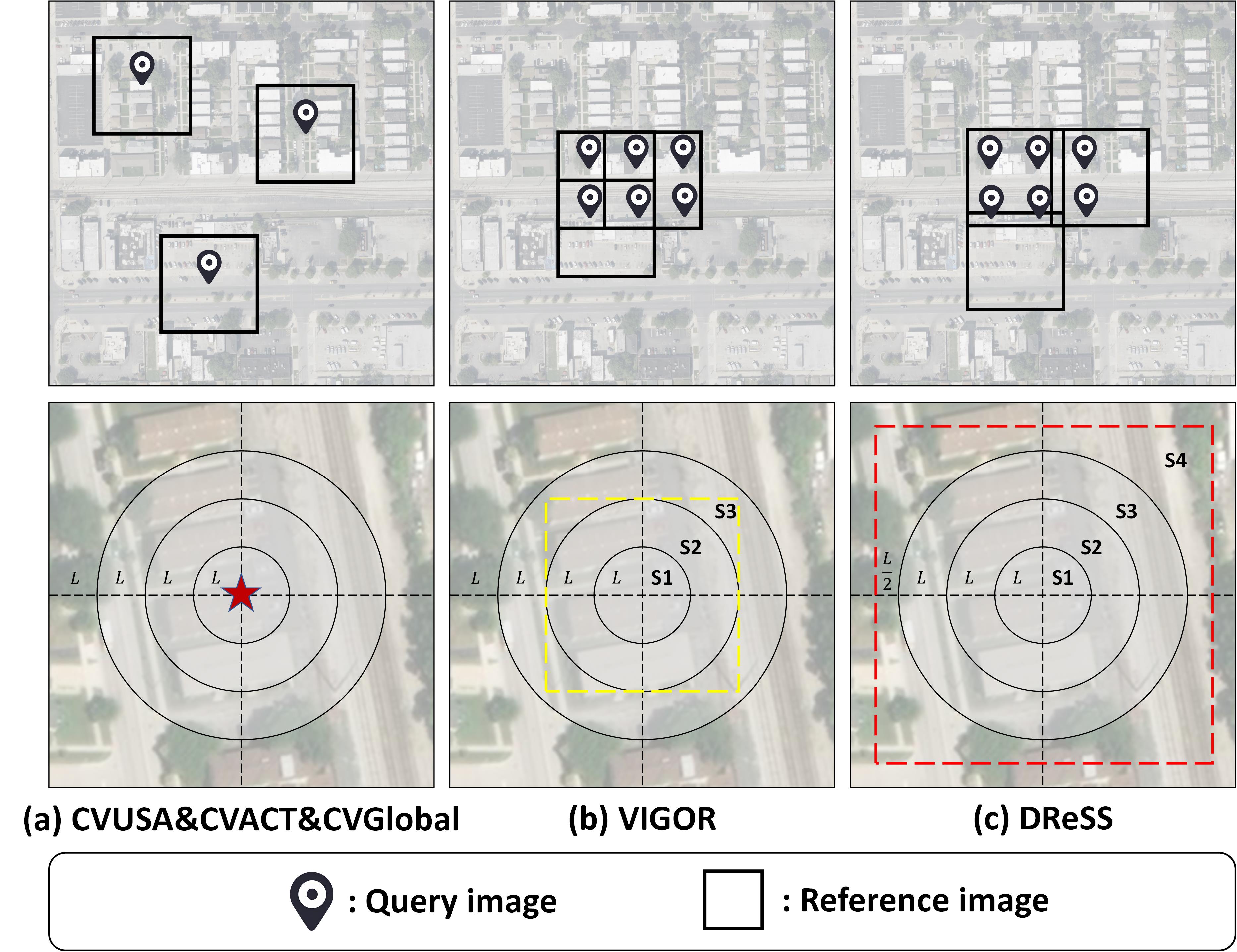}
    \caption{Visualization of decentrality conditions across different datasets. The red star in CVUSA, CVACT and CVGlobal represents center alignment, while the yellow box in VIGOR and the red box in DReSS represent the hit areas, indicating different degrees of decentrality.}
    \label{fig:vis_decentrality}
\end{figure}

Table \ref{tab:head-to-head comparison} provides a detailed comparison between our dataset and previous datasets CVUSA \citep{zhai2017predicting}, CVACT \citep{liu2019lending}, VIGOR \citep{zhu2021vigor} and CVGlobal \citep{ye2025cross}. We also visualize the decentrality conditions in existing datasets alongside our proposed DReSS dataset for comparison, as shown in Figure \ref{fig:vis_decentrality}. In the CVUSA \citep{zhai2017predicting}, CVACT \citep{liu2019lending} and CVGlobal \citep{ye2025cross} datasets, decentrality is overlooked due to the simple center-aligned settings, which are inadequate for real-world applications where perfectly matched reference images cannot be guaranteed. In the VIGOR \citep{zhu2021vigor} dataset, due to a high overlap of 50\% and dataset balancing, the best-matched image pairs are mostly confined to the yellow box, indicating a limited portion of decentrality. 

The statistical analysis results of VIGOR are also presented in Table \ref{tab:subsets_vigor}. While VIGOR attempts to define a hit area for positive samples, a small proportion of the best-matched image pairs (positive samples) are erroneously assigned to subsets with high decentrality. Additionally, although the dataset includes "semi-positive" street-view images with larger decentrality, there is always an alternative reference image available to establish a "positive" pair for each street-view image. Consequently, the VIGOR dataset is inadequate for effectively investigating the decentrality issue, even with a reorganization of its original structure.

\begin{table}
    \centering
    \begin{tabular}{cc|cccc}
    \toprule
    \toprule
         Overlap&  Subset&  S1&  S2&  S3& S4\\
         \midrule
         50\%&  Train
&  9,037&  29,324&  14,137& 111\\
         VIGOR&  Test&  9,001&  29,544&  13,957& 103\\
         \bottomrule
         \bottomrule
    \end{tabular}
    \caption{Number of best-matched cross-view image pairs in four subsets in VIGOR dataset.}
    \label{tab:subsets_vigor}
\end{table}

\begin{figure}[h!]
    \centering
    \includegraphics[width=1.0\linewidth]{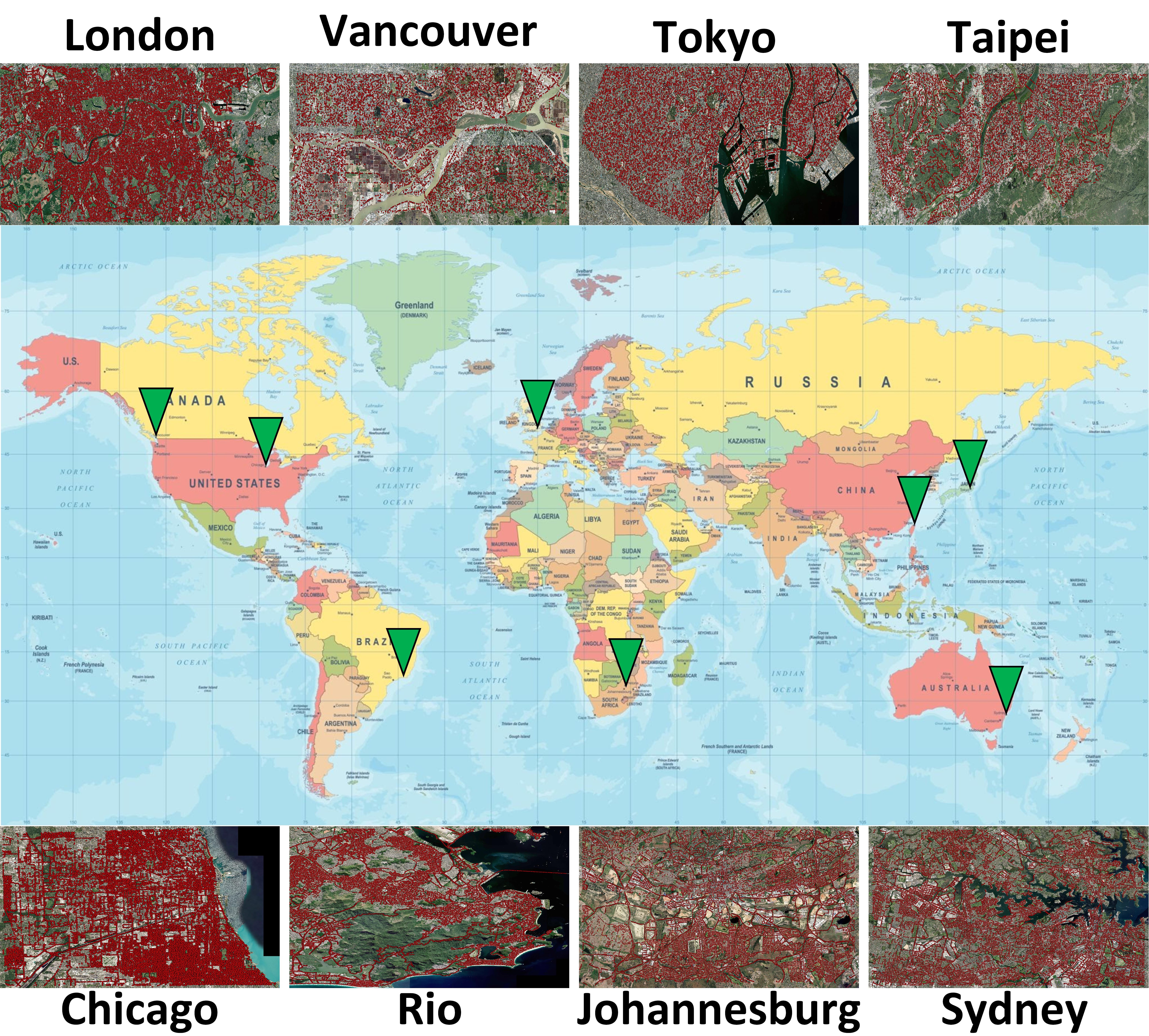}
    \caption{Aerial images of eight cities with diverse landscapes from across the world and the distributions of panoramas (red dots) in the DReSS dataset.}
    \label{fig:dataset_location}
\end{figure}

Compared to previous datasets, our proposed DReSS dataset can better cover the issue of decentrality, providing a more comprehensive testbed for CVGL. In addition to introducing the issue of decentrality, DReSS covers a larger scope and a wider variety of landscapes. As illustrated in Figure \ref{fig:dataset_setting}(b) and Figure \ref{fig:dataset_location}, DReSS encompasses urban and suburban areas. For example, the coverage of Chicago in DReSS (619 \(km^2\)) is almost 20 times larger than in VIGOR (30.3 \(km^2\)). DReSS includes samples from eight cities worldwide, each exceeding 400 \(km^2\). This makes DReSS the largest and most realistic dataset for cross-view geo-localization.

\subsection{Data Collection} 
As shown in Figure \ref{fig:dataset_location}, the DReSS dataset covers over 400 \(km^2\) in each of eight diverse cities around the world. The dataset consists of 422,760 aerial images sourced from Esri World Imagery \citep{aerialview}, captured at zoom level 18 with a ground resolution of approximately 0.597 \(m\). Each aerial image has a resolution of \(224\times 224\) pixels. Additionally, DReSS features 174,934 street-view panoramas obtained using the Google Street View. These panoramas are randomly distributed within the coverage area of the aerial images, with an average interval of about 500 \(m\) between samples. The panoramas are North-aligned, and each has a resolution of \(2048\times 1024\) pixels. To comply with Google Street View's policies, we will provide only the panorama IDs and instructions for downloading them, enabling users to access the dataset independently.

\subsection{The Evaluation Protocol} 
We adopt the SAME and CROSS settings as used in VIGOR. In the SAME, the model is trained with training data from all cities and tested with testing data from all cities. In the CROSS setting, the model is trained with data from four cities (Chicago, London, Sydney, Tokyo) and tested with data from the remaining four cities (Johannesburg, Rio, Taipei, Vancouver). Furthermore, a novel evaluation is proposed within the testing data of DReSS to assess the model's robustness for cross-view geo-localization with increasing decentrality.

\section{Methodology}
\label{Methodology}

\subsection{Problem Formulation}
Denote a set of cross-view street-aerial image pairs as \(\left \{  (I^s_i,I^a_i)\right \} ^N\), where the  \(I^s\) and \(I^a\) represent street-view and aerial-view images, respectively. \(N\) represents the number of pairs. The cross-view images in each pair represent the same geo-location, with each pair corresponding to a different geo-location. In cross-view geo-localization, given a query street image \(I_q^s\), the goal is to retrieve the best matching reference aerial image \(I_r^a, r\in \left \{  1,2,\dots,N\right \} \) with geo-tag, to determine the geo-location of query image \(I_q^s\).

For a given cross-view image set, \(\left \{  (I^s_i,I^a_i)\right \} ^N\), we denote the image representations generated by the encoder as \(\left \{  (f^s_i,f^a_i)\right \} ^N\). These representations must possess the attribute that the similarity between the representations of a matched image pair is higher than that between unmatched pairs. Consequently, denote the cosine similarity function as \(sim(\cdot,\cdot )\), the cross-view geo-localization task can be formulated as:
\[
r = \underset{i \in \{1, \ldots, N\}}{\arg\max} \, sim(f_q^s, f_i^a) \tag{1}.
\]
\indent If the retrieved result is right, \(r\) equals to \(q\). To keep the notation straightforward, we will leave out the subscript \(i\) in the next sections, except when we discuss the loss function.

\begin{figure*}[h!]
    \centering
    \includegraphics[width=1.0\linewidth]{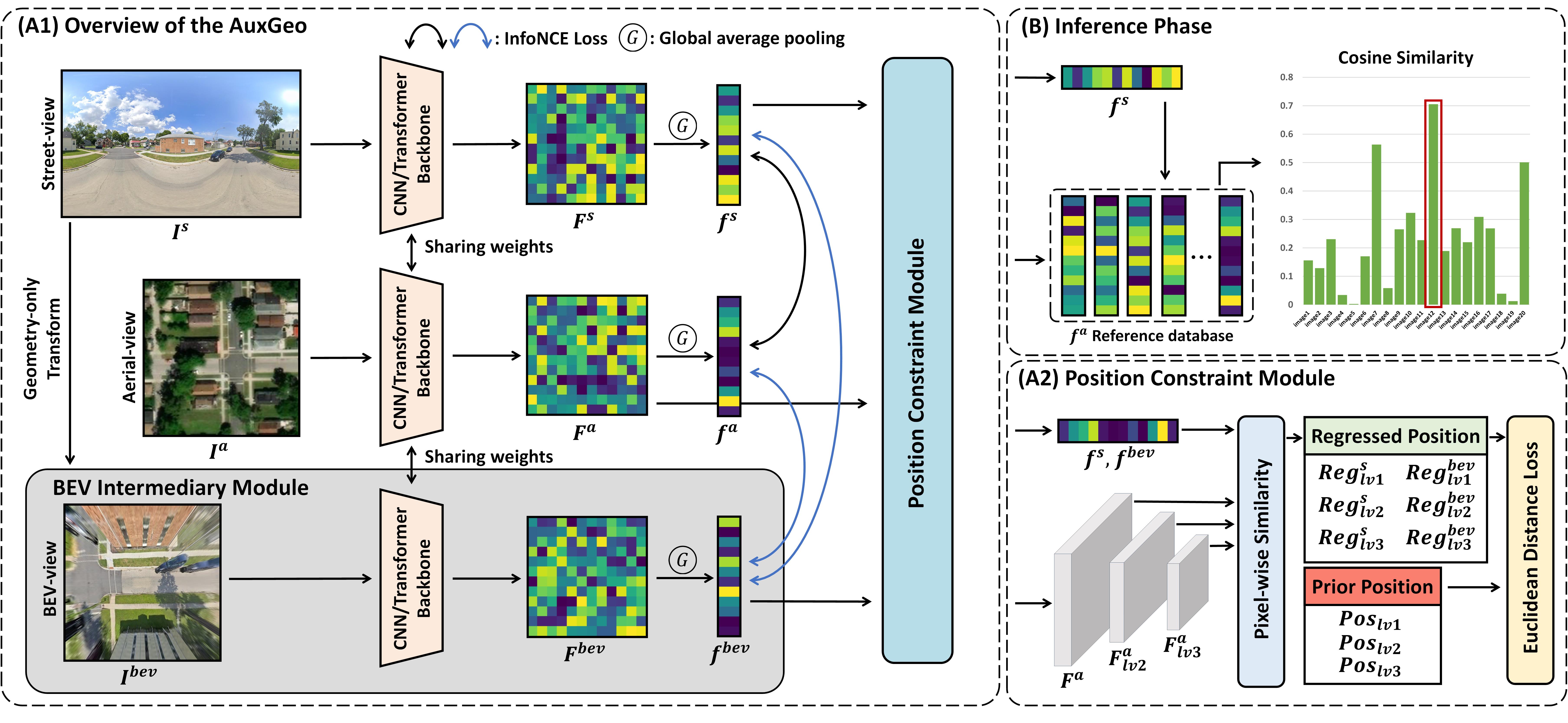}
    \caption{(A1) Overview of our proposed method AuxGeo with two novel modules BIM and PCM. (B) Illustration of the inference phase of the AuxGeo, which demonstrates that the proposed modules act as components of the multi-metric optimization and take no extra cost during inference. (A2) Illustration of the proposed PCM module.}
    \label{fig:framework}
\end{figure*}

\subsection{AuxGeo Model}
Given a cross-view image pair \((I^s, I^a)\), the fine-grained features generated by the backbone are denoted as \((F^s, F^a)\), where \(F^s \in \mathbb{R}^{H^s \times W^s \times C}\) and \(F^a \in \mathbb{R}^{H^a \times W^a \times C}\). Here, \(H\), \(W\), and \(C\) represent height, width, and channels, respectively. Existing methods achieve superior representations \((f^s, f^a)\), where \(f^s, f^a \in \mathbb{R}^{C}\), from the fine-grained features using self-attention \citep{zhu2022transgeo,zhu2023simple} or feature recombination \citep{zhang2024aligning}. However, it has been demonstrated that these capabilities can be replaced by a robust backbone with a simple global average pooling \citep{deuser2023sample4geo}. Therefore, moving beyond current methods, our proposed AuxGeo focuses on leveraging information that cannot be directly included in the representation but can enhance the backbone's capabilities. 

\subsubsection{Model Overview}
The proposed AuxGeo incorporates a multi-metric optimization network, which includes a contrastive representation learning structure derived from Sample4Geo \citep{deuser2023sample4geo} and two auxiliary modules, as depicted in Figure \ref{fig:framework}(A1). 

During training, the BEV Intermediary Module (BIM) establishes an additional branch using the BEV-view image \(I^{bev}\) generated from \(I^s\) with a geometry-only transform. The BEV-view branch shares the same backbone as the Street and Aerial-view branches, acting as an intermediary to address the cross-view problem with decentrality. Subsequently, \(f^s\), \(f^{bev}\), and \(F^a\) are processed through the Position Constraint Module (PCM), which mitigates the decentrality issue by imposing additional constraints in a multi-grained fashion. During inference, as illustrated in Figure \ref{fig:framework}(B), the generation of representations requires only a straightforward process, incurring no additional pre-processes or costs. In the ensuing sections, we provide a detailed description of the core modules of AuxGeo.

\subsubsection{BEV Intermediary Module} 
The BIM (BEV Intermediary Module) uses the BEV-view image \(I^{bev}\) generated from the street-view panorama \(I^s\) through a geometry-only transformation introduced by Wang et al. \citep{wang2024fine} as its input. While the BEV image effectively overcomes the cross-view problem under decentrality conditions, it only retains ground-level details (like road markings) and omits others (such as houses and trees) because it’s a subset of the street-view panorama. Therefore, we incorporate BEV-view images as intermediaries in our method rather than using them directly as query images.

\begin{figure}[h!]
    \centering
    \includegraphics[width=0.9\linewidth]{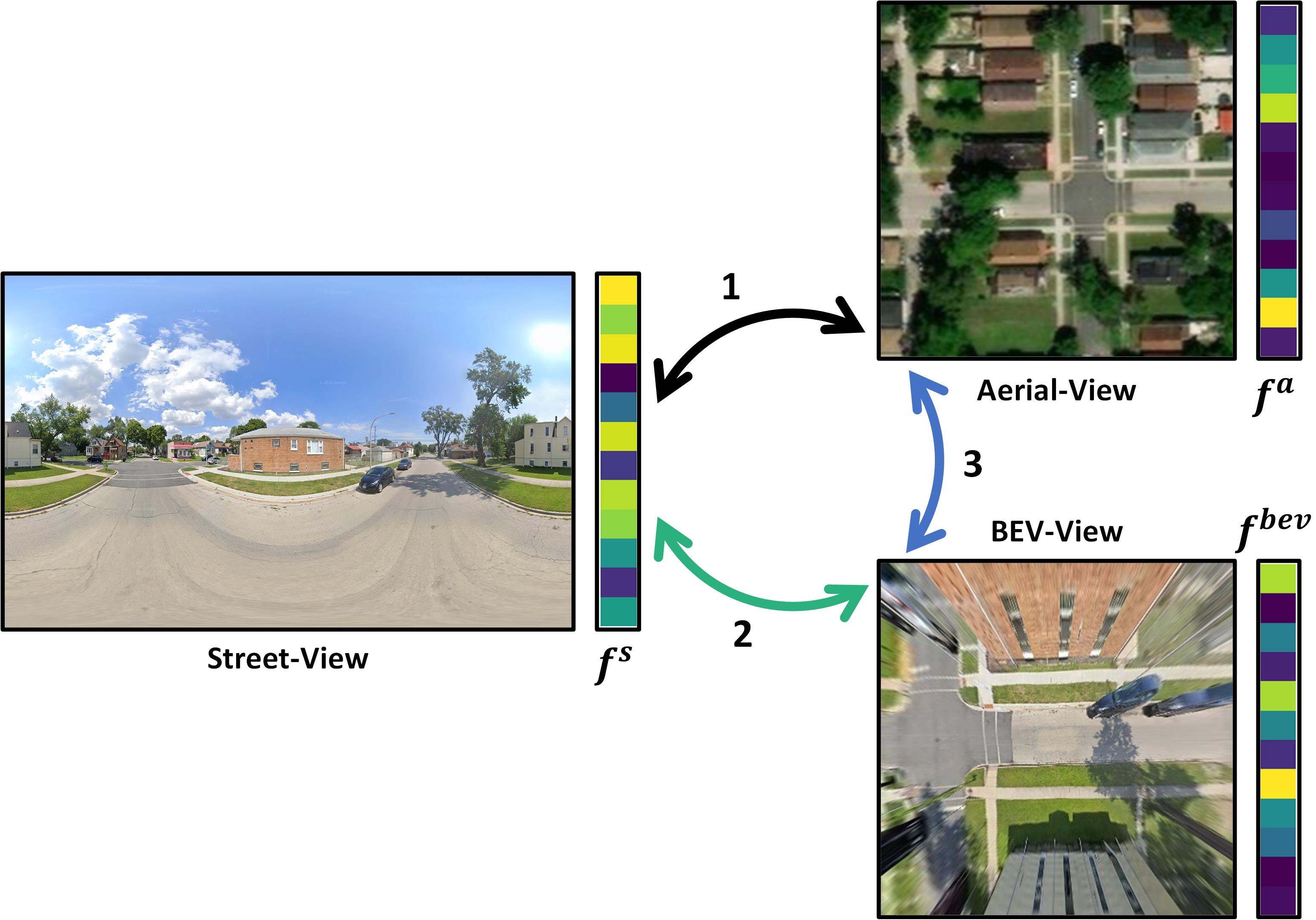}
    \caption{The process of the BEV Intermediary Module (BIM). The black double-arrow 1 indicates the original cross-view matching with decentrality. With the incorporation of the BEV image, the problem is divided into: cross-view matching between the street-view and BEV-view (double-arrow 2), and decentrality adjustment (double-arrow 3) between the aerial-view and BEV-view.}
    \label{fig:BIM_fig}
\end{figure}

In the BIM, the BEV image is integrated into the contrastive representation learning process. The BEV-view mitigates the cross-view problem with decentrality by introducing only a cross-view problem between the street-view and BEV-view and only a decentrality problem between the aerial-view and BEV-view, as depicted in Figure \ref{fig:BIM_fig}. 

In addition to the original constraint (represented by the black double arrow in Figure \ref{fig:BIM_fig} between street and aerial views, the BIM introduces new constraints (represented by the green and blue double arrows in Figure \ref{fig:BIM_fig}. We utilize the symmetric InfoNCE loss as the loss function to supervise the training, as indicated in Equation (2): 
\[\mathcal{L}_{InfoNCE}(f^s_q,\left \{ f^a_i \right \}^N )=-log\frac{exp(f_q^s\cdot f_+  ^a/\tau )}{ {\textstyle \sum_{i=0}^{N}}exp(f^s_q\cdot f^a_i/\tau ) }, \tag{2}\]
where the \(f_+  ^a\) is the matched representation, and the temperature \(\tau\) is a hyper-parameter that can be learnable or static.

\subsubsection{Position Constraint Module} 
The Position Constraint Module (PCM) processes the coarse-grained features \(f^s\) and \(f^{bev}\) alongside the fine-grained feature \(F^a\), using position prior information as constraints to enhance the backbone's performance involving decentrality, as depicted in Figure \ref{fig:framework}(A2). The position prior \(Pos\), indicating the location of the query image on the reference image, can be easily determined using geo-tags. Unlike previous offset regression methods that apply the position prior only at the coarse-grained level \citep{zhu2021vigor}, we recognize that the aerial-view image, acting as the map, typically has a larger coverage compared to the street-view image as the query. Therefore, we designed a multi-grained position regression task within the PCM. 

\begin{figure}[h!]
    \centering
    \includegraphics[width=1.0\linewidth]{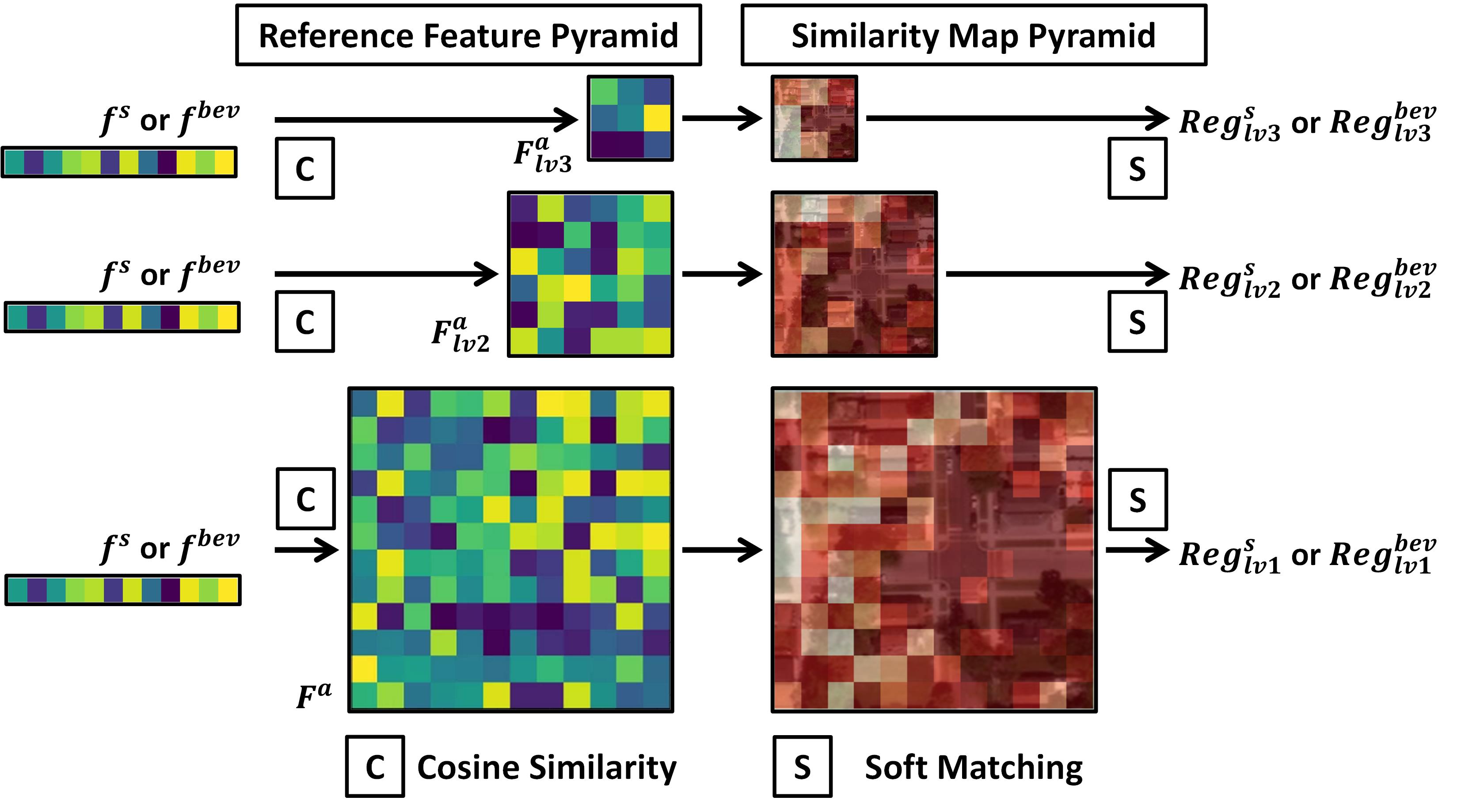}
    \caption{The process of the Position Constraint Module (PCM). Similarity calculations produce multi-level pyramid similarity maps, from which regressed positions are derived through a soft matching process.}
    \label{fig:PCM_fig}
\end{figure}

The coarse-grained features \(f^s\) and \(f^{bev}\) are utilized to compute the similarity with each pixel of the fine-grained feature \(F^a\), producing similarity maps \(M^{s}\) and \(M^{bev}\). This process does not involve any additional learnable components to maintain the backbone's representation purity. From these similarity maps, the regressed positions \(Reg^s\) and \(Reg^{bev}\) are derived through a soft matching process, as depicted in Figure \ref{fig:PCM_fig}. Considering the variation in coverage ratio between cross-view images in different scenarios (for example, the aerial-view image in DReSS has a larger coverage than that in VIGOR, while the street-view image always has a similar coverage), a feature pyramid is employed in the PCM to utilize the position prior at different scales. The original fine-grained feature is processed at multiple scales: \(lv1\) for the original feature, \(lv2\) for 2x down-sampled features, and \(lv3\) for 4x down-sampled features, each generated using average pooling. In this setup, position priors at different scales are denoted as \(Pos_{lvi}\), and the regressed positions at different levels are denoted as \(Reg^s_{lvi}\) and \(Reg^{bev}_{lvi}\) , obtained through a soft matching process from the similarity maps. To supervise the network during training, an L2 Euclidean distance loss, denoted as \(d(\cdot, \cdot)\), is employed. \(N\) represents the number of layers in the feature pyramid, and \(N=3\). The loss function for the PCM is expressed as follows:
\[
\mathcal{L}_{PCM} = \sum_{i=1}^{N} \left( d(Pos_{lvi}, Reg^s_{lvi}) + d(Pos_{lvi}, Reg^{bev}_{lvi}) \right). \tag{3}
\]
\indent Equation (3) encapsulates the multi-level loss function employed to train the PCM, ensuring that position constraints are effectively utilized across multiple granularities.

\subsubsection{Loss Function} 
During training, the overall loss function is denoted as \(\mathcal{L}oss\), as defined below:
\[
\mathcal{L}oss = \mathcal{L}^{street}_{aerial} + \lambda_1 \cdot \mathcal{L}^{bev}_{street} + \lambda_2 \cdot \mathcal{L}^{bev}_{aerial} + \lambda_3 \cdot \mathcal{L}_{PCM}, \tag{4}
\]
where \(\mathcal{L}_{b}^{a}\) represents the InfoNCE loss between \(a\) and \(b\) views. \(\lambda_1\), \(\lambda_2\), and \(\lambda_3\) are hyperparameters for balancing the weights of the losses.

\section{Experiments}
\label{Experiments}

\subsection{Datasets and Experimental Settings}
\subsubsection{Datasets} 
We evaluate AuxGeo on four cross-view geo-localization datasets: CVUSA, CVACT, VIGOR, and our proposed DReSS dataset. CVUSA and CVACT employ a center-aligned, one-to-one retrieval paradigm between cross-view images, whereas VIGOR and DReSS extend beyond one-to-one retrieval, featuring non-center-aligned settings. Notably, DReSS supports cross-view geo-localization in a significantly larger and seamless environment, encompassing diverse styles of cities worldwide.
\begin{itemize}
\item CVUSA includes 35,532 image pairs for training and 8,884 image pairs for testing, with images primarily collected from suburban areas.
\item CVACT provides 35,532 image pairs for training and 8,884 image pairs for validation (CVACT\_val), along with an additional 92,802 image pairs for city-scale geo-localization testing (CVACT\_test). The images are densely collected from the urban area of Canberra.
\item VIGOR consists of 105,214 street images and 90,618 aerial images from four cities in the US. The reference images cover the area seamlessly without center-aligned settings in cross-view images, primarily collected from urban areas.
\item DReSS includes 174,934 street images and 422,760 aerial images from eight cities worldwide. The reference images provide a much larger and seamless coverage, without center-aligned settings. The dataset spans urban, suburban, and rural areas.
\end{itemize}

\subsubsection{Evaluation Metrics} 
Following previous works, we use the R@\(K,K=\left \{ {1,5,10,1\%} \right \} \) as the evaluation metric, which represents the ratio of correct retrievals within the top \(K\) results. Additionally, for VIGOR and DReSS, we report the hit rate, indicating the probability that the top-1 retrieved reference image covers the query image.

\subsubsection{Implementation Details} 
We employ a ConvNeXt-B backbone \citep{liu2022convnet} pretrained on ImageNet-22K. Following previous works, we use an input resolution of \(384 \times 768\) for street-view images and \(384 \times 384\) for aerial-view images. In Equation (4), the hyperparameters \(\lambda_1\), \(\lambda_2\), and \(\lambda_3\) are set to 0.1, 0.1, and 0.05, respectively. Models are trained on a server equipped with 8 NVIDIA V100-32G GPUs, using a batchsize of 128.

\subsection{Comparison with State-of-the-art Methods}
We comapre our AuxGeo with several state-of-the-art methods, including TransGeo \citep{zhu2022transgeo}, GeoDTR \citep{zhang2023cross}, FRGeo \citep{zhang2024aligning} and our baseline Sample4Geo \citep{deuser2023sample4geo}.

\begin{table*}[h]
    \centering
    \resizebox{\textwidth}{!}{
    \begin{tabular}{l|c|ccccc|ccccc}
        \toprule
    \toprule
 Method& Backbone
&& \multicolumn{3}{c}{DReSS}& && \multicolumn{3}{c}{VIGOR}&\\
         & &R@1&  R@5&  R@10&  R@1\%& Hit &R@1& R@5& R@10& R@1\%&Hit \\
    \midrule
         \textbf{SAME}& &&  &  &  &  && & & &\\
         TransGeo& {\small Deit-small}&19.86&  40.29&  49.39&  98.60& --&61.48& 87.54& 91.88& 99.56&73.09 \\
         FRGeo& *&27.19&  46.17&  53.82&  98.01& --
&71.26& 91.38& 94.32& 99.52&82.41 
\\
         Sample4Geo& {\small ConvNeXt-B}&51.40&  72.78&  78.52&  98.29& 55.89
&77.86& 95.66& 97.21& 99.61&93.64
\\
         Ours& {\small ConvNeXt-B}&\textbf{54.70}&  \textbf{76.09}&  \textbf{81.46}&  \textbf{98.74}& \textbf{59.20}&\textbf{80.34}& \textbf{96.25}& \textbf{97.57}& \textbf{99.67}&\textbf{93.78}\\
    \midrule
         \textbf{CROSS}& &&  &  &  &  && & & &\\
         TransGeo& {\small Deit-small}&4.09&  11.45&  16.26&  77.86& --&18.99& 38.24& 46.91& 88.94&21.21 \\
         FRGeo& *&8.07&  18.23&  24.07&  82.43& --
&37.54& 59.58& 67.34& 94.28&40.66 
\\
         Sample4Geo& {\small ConvNeXt-B}&28.23&  48.19&  56.03&  92.93& 30.65&61.70& 83.50& 88.00& \textbf{98.17}&74.78
\\
         Ours& {\small ConvNeXt-B}&\textbf{32.44}&  \textbf{53.18}&  \textbf{60.98}&  \textbf{93.82}& \textbf{34.90}&\textbf{63.94}& \textbf{84.98}& \textbf{88.98}& 98.02&\textbf{76.25} \\
    \bottomrule
    \bottomrule
    \end{tabular}}
    \caption{Quantitative comparison between our AuxGeo and current state-of-the-art methods on VIGOR and DReSS. * indicates that the backbone used for FRGeo on DReSS is ConvNeXt-B for fair comparison, while on VIGOR, ConvNeXt-T is used as reported in the original paper. The best results are shown in bold.}
    \label{tab:results_vigor_DReSS}
\end{table*}

\subsubsection{Results on VIGOR and DReSS} 
Beyond the center-aligned setting, we report the cross-view geo-localization performance on VIGOR and DReSS, with results shown in Table \ref{tab:results_vigor_DReSS}. On VIGOR, AuxGeo achieves state-of-the-art performance. This improvement is attributed to the enhanced capability of the backbone network, activated by the BIM and PCM modules in AuxGeo. On DReSS, performance significantly decreases due to the greater decentrality. Despite this, AuxGeo can also outperform previous state-of-the-art methods, improving the R@1 by 3.30\% in SAME and 4.21\% in CROSS compared to the previous state-of-the-art method, Sample4Geo.

\begin{table}[h]
    \centering
    \begin{tabular}{l|cccc}
    \toprule
    \toprule
         {\small Method}&  {\small Subset 1}&  {\small Subset 2}&  {\small Subset 3}& {\small Subset 4}\\
    \midrule
         {\small Sample4Geo}
&  72.74&  68.30&  57.46& 35.49\\
         {\small Ours}&  \textbf{74.42} &  \textbf{70.06}&  \textbf{61.08}& \textbf{39.50}\\
    \midrule
 {\small Improvement}& 1.68 \(\uparrow \)& 1.76 \(\uparrow \)& 3.62 \(\uparrow \)&4.01 \(\uparrow \)\\
 \bottomrule
    \bottomrule
    \end{tabular}
    \caption{Quantitative comparison of R@1 score on subsets of DReSS SAME, considering increasing degrees of decentrality. The best results are shown in bold.}
    \label{tab:results_decentrality}
\end{table}

\subsubsection{Results on Increasing Decentrality} 
To further validate the effectiveness of mitigating the decentrality problem, we conducted a series of experiments. Models were trained using data with various degrees of decentrality and evaluated across Subsets 1 to 4 within the testing data. The results, presented in Table \ref{tab:results_decentrality}, reveal the following: a) Geo-localization performance significantly degrades with increasing decentrality. The performance of both the Sample4Geo and AuxGeo on Subsets 1 and 2 is comparable to that on the VIGOR dataset, where decentrality is similar to that of Subsets 1 and 2 in DReSS. However, performance on Subsets 3 and 4 experiences a sharp decline for both methods, indicating that decentrality poses a substantial challenge, warranting further investigation. b) AuxGeo, despite experiencing a performance decrease, demonstrates progressively better improvement compared to the baseline. This suggests that the proposed modules effectively contribute to overcoming the challenges induced by decentrality.

\begin{table}
    \centering
    \begin{tabular}{l|ccccc}
    \toprule
    \toprule
         {\small Method}&  {\small R@1}&  {\small R@5}&  {\small R@10}&  {\small R@1\%}& {\small Hit}\\
         \midrule
         \textbf{SAME}&  \multicolumn{5}{c}{VIGOR \(\longrightarrow \) DReSS}\\
         {\small Sample4Geo}&  8.93&  16.21&  19.99&  62.12& 9.16\\
         {\small Ours}&  \textbf{9.24}&  \textbf{16.69}&  \textbf{20.57}&  \textbf{63.20}& \textbf{9.46}\\
         \midrule
         \textbf{SAME}&  \multicolumn{5}{c}{DReSS \(\longrightarrow \) VIGOR}\\
         {\small Sample4Geo}&  22.39&  47.65&  57.13&  92.87& 38.78\\
         {\small Ours}&  \textbf{30.46}&  \textbf{59.42}&  \textbf{68.42}&  \textbf{95.55}& \textbf{50.06}\\
         \bottomrule
         \bottomrule
    \end{tabular}
    \caption{Generalization performance when trained on VIGOR and evaluated on DReSS and vice versa. The best results are shown in bold.}
    \label{tab:generalization}
\end{table}

\subsubsection{Results on Generalization Capabilities} 
Besides the evaluation of the CROSS settings in VIGOR and DReSS, we also evaluated the generalization capabilities of models trained on VIGOR and tested on DReSS, and vice versa, to test the transferability between images from different sources. The results, shown in Table \ref{tab:generalization}, demonstrate that AuxGeo achieves better performance. This advantage is particularly emphasized in the generalization capabilities from larger datasets (DReSS) to smaller ones (VIGOR), which indicates the importance of the auxillary contents provided by modules in AuxGeo.
\begin{table*}[h]
    \centering
    \resizebox{\textwidth}{!}{
    \begin{tabular}{l|cccc|cccc|cccc}
    \toprule
    \toprule
         Method&  &\multicolumn{2}{c}{CVUSA}&&  &\multicolumn{2}{c}{CVACT Val}&&  &\multicolumn{2}{c}{CVACT Test}&\\
         \midrule
         &  R@1&  R@5&  R@10&  R@1\%&  R@1&  R@5&  R@10&  R@1\%&  R@1&  R@5&  R@10& R@1\%\\
         TransGeo&  94.08&  98.36&  99.04&  99.77&  84.95&  94.14&  95.78&  98.37&  -&  -&  -& -\\
 GeoDTR\dag{}& 95.43& 98.86& 99.34& 99.86& 86.21& 95.44& 96.72& 98.77& 64.52& 88.59& 91.96&98.74\\
         FRGeo&  97.06&  99.25&  99.47&  99.85&  90.35&  96.45&  97.25&  98.74&  72.15&  91.93&  94.05& 98.66\\
         Sample4Geo&  98.68&  99.68&  \textbf{99.78}&  \textbf{99.87}&  90.81&  96.74&  97.48&  98.77&  71.51&  92.42&  94.45& 98.70\\
         Ours&  \textbf{98.80}&  \textbf{99.71}&  99.75&  99.85&  \textbf{91.86}&  \textbf{97.23}&  \textbf{97.79}&  \textbf{98.93}&  \textbf{73.65}&  \textbf{93.54}&  \textbf{95.23}& \textbf{98.75}\\
    \bottomrule
    \bottomrule
    \end{tabular}}
    \caption{Quantitative comparison between our AuxGeo and current state-of-the-art methods on CVUSA and CVACT. \dag{}  indicates applying the polar transform to reference image. The best results are shown in bold.}
    \label{tab:results_cvusa&cvact}
\end{table*}

\subsubsection{Results on CVUSA and CVACT} The results of cross-view geo-localization on CVUSA and CVACT are presented in Table \ref{tab:results_cvusa&cvact}. Compared to previous works, our findings indicate that AuxGeo achieves the best performance, demonstrating that polar transform or feature recombination is unnecessary during inference. Additionally, the extra features provided by our modules still improve performance in center-aligned settings, highlighting their effectiveness.

\section{Discussion}
\label{Discussion}

\subsection{Effectiveness of Components} 
To evaluate the effectiveness of the proposed modules (BIM and PCM) in the AuxGeo method, we conducted a series of ablation experiments. For fair comparison, the settings of strategies during training and testing remained consistent across all configurations. The results on VIGOR and DReSS datasets are presented in Table \ref{tab:results_ablation_study}. The results demonstrate that integrating either BIM or PCM yields improvements, with the best performance achieved when both BIM and PCM are incorporated. These findings offer a new perspective on cross-view geo-localization. As an image retrieval-based method, it demonstrates that certain elements, which can't be directly used during inference (such as partial viewpoint transform and fine-grained cross-view feature correspondences), can be effectively utilized as auxiliary components during training. This approach maximizes the use of these elements to better train the backbone model.

\begin{table}[t]
    \centering
     \begin{tabular}{l|ccccc}
    \toprule
    \toprule
         {\small Method}&  {\small R@1}&  {\small R@5}&  {\small R@10}&  {\small R@1\%}& {\small Hit}\\
         \midrule
 & \multicolumn{5}{c}{VIGOR SAME}\\
         \midrule
         {\small Baseline}&  77.86&  95.66&  97.21&  99.61& 93.64\\
         {\small w/ BIM}&  78.72&  96.29&  97.55&  99.67& 94.25\\
         {\small w/ PCM}&  79.88&  96.01&  97.39&  99.66& 93.47\\
         {\small w/ BIM, PCM}&  80.34&  96.25&  97.57&  99.67& 93.78\\
         \midrule
         &  &  \multicolumn{3}{c}{DReSS SAME}& \\
         \midrule
         {\small Baseline}&  51.40&  72.78&  78.52&  98.29& 55.89\\
         {\small w/ BIM}&  54.12&  75.62&  81.07&  98.61& 58.82\\
 {\small w/ PCM}& 53.12& 74.35& 80.00& 98.69&57.05\\
 {\small w/ BIM, PCM} & 54.70& 76.09& 81.46& 98.74&59.20\\
 \bottomrule
 \bottomrule
    \end{tabular}
    \caption{Ablation study of the effectiveness of the proposed modules. Our baseline model is the Sample4Geo.}
    \label{tab:results_ablation_study}
\end{table}

\begin{table}[h]
    \centering
    \begin{tabular}{l|cccc}
    \toprule
    \toprule
 & \multicolumn{4}{c}{VIGOR Same-area}\\
         &  R@1&  R@5&  R@10&  R@1\%\\
         \midrule
 Pano only (baseline)& 77.86& 95.66& 97.21&99.61\\
 BEV only& 38.05& 61.73& 69.33& 95.08\\
 BEV+Pano& 76.17& 94.72& 96.50& 99.60\\
         Pano (BIM)&  78.72&  96.29&  97.55&  99.67\\
         \bottomrule
         \bottomrule
    \end{tabular}
    \caption{Quantitative comparison of the strategies for utilizing the BEV image.}
    \label{bev-strategies}
\end{table}

\begin{figure}[h!]
    \centering
    \includegraphics[width=1\linewidth]{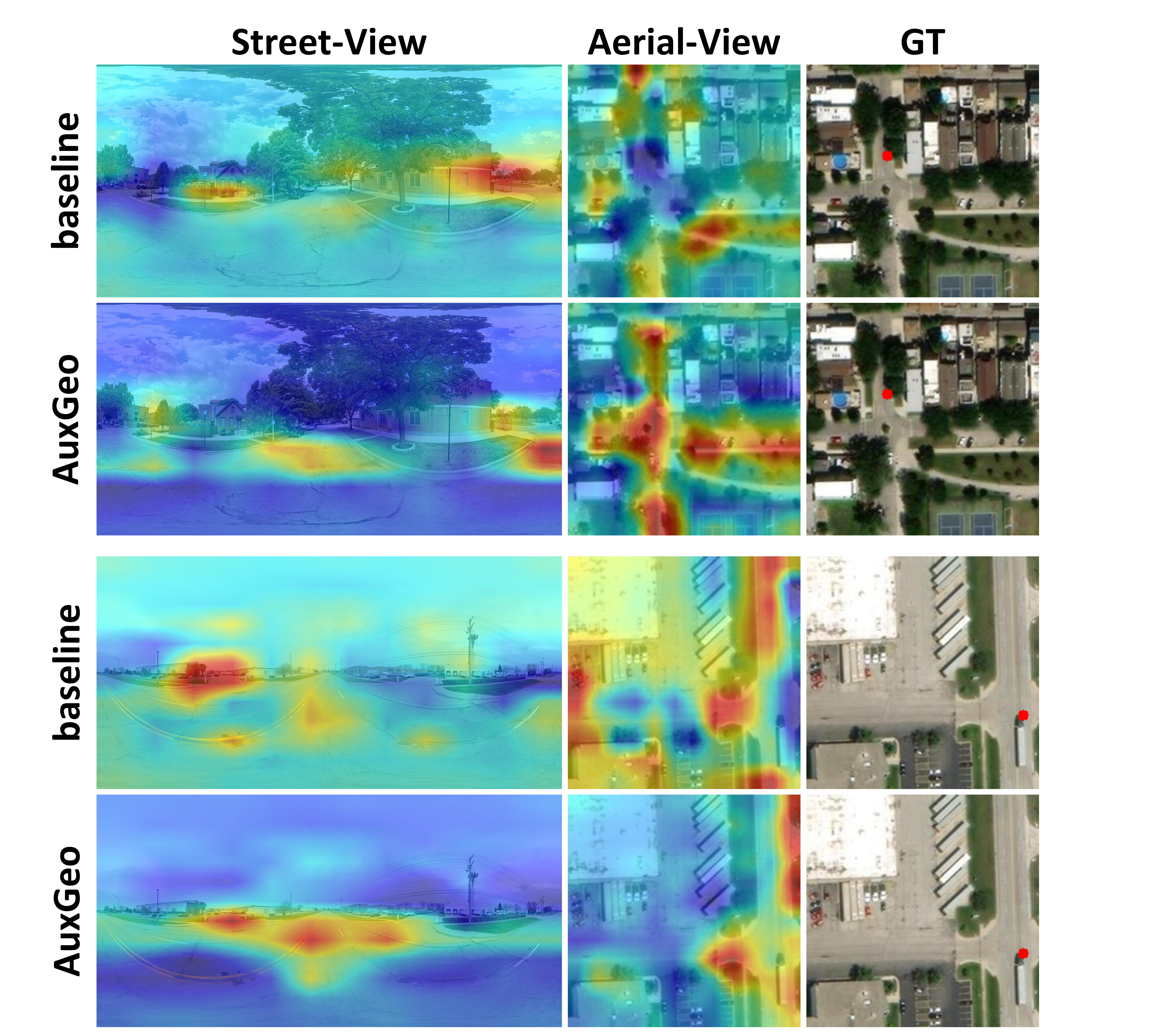}
    \caption{Heatmap visualization of DReSS dataset images generated by baseline (Sample4Geo) and AuxGeo models. The third column indicates the ground truth position (red dot) of the street-view image on the corresponding aerial-view image.}
    \label{fig:visualization}
\end{figure}

\begin{figure}[h!]
    \centering
    \includegraphics[width=1\linewidth]{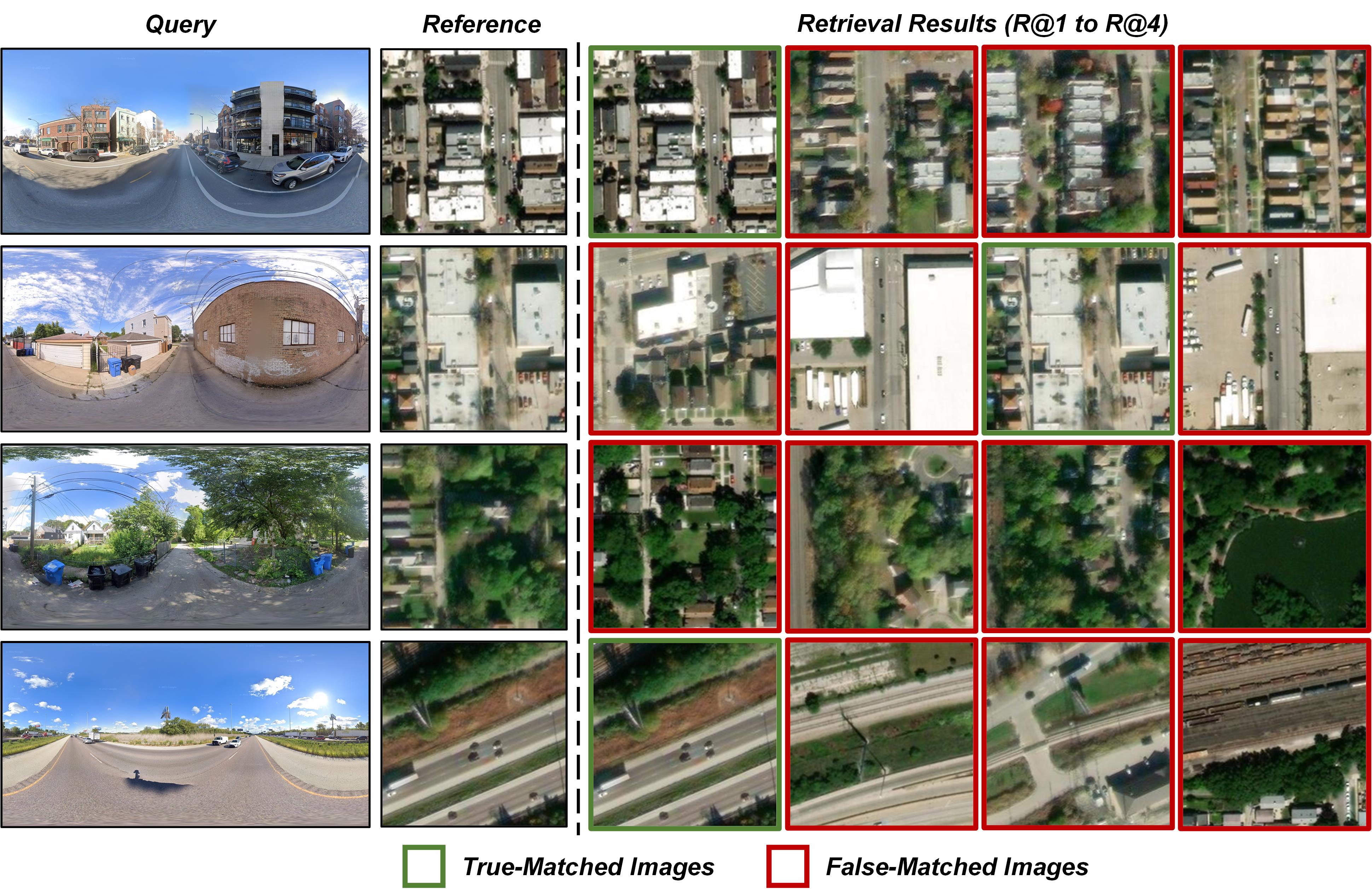}
    \caption{Visualization results of the cross-view geo-localization in DReSS dataset.}
    \label{fig:visualization_retrieval}
\end{figure}

\subsection{Utilization Strategy for the BEV image}

To identify the optimal utilization strategy for the proposed BEV Intermediary Module (BIM), a comparative study was conducted focusing on the use of BEV images. Four configurations were evaluated: (a) using only the panorama as the query (baseline); (b) using only the BEV image as the query; (c) summing features from both BEV and panorama images; and (d) employing the BEV image as an intermediary (BIM). All models were trained under identical conditions to ensure a fair and consistent comparison. The results are presented in Table \ref{bev-strategies}. Although the BEV transform substantially mitigates viewpoint variations under the decentrality problem, it causes a significant loss of visual information from the original image, leading to decreased performance. Moreover, when comparing the use of only the panorama or adding the features of the BEV and panorama images, our BIM configuration delivers the highest performance. Notably, BIM achieves this without incurring additional costs beyond the baseline, whereas the feature summation approach requires additional processing and computational resources.

Based on these findings, we propose BIM, which leverages the BEV image as an intermediary to achieve performance improvements efficiently.

\subsection{Visualization Analysis} 
To qualitatively evaluate AuxGeo's effectiveness, we visualize heatmaps of features from the baseline and AuxGeo models using images from the test set of DReSS. As shown in Figure \ref{fig:visualization}, the upper two rows depict a narrow neighborhood scene, while the lower two rows depict an open crossroads scene. The heatmaps reveal that, compared to the baseline (Sample4Geo), AuxGeo more effectively focuses on salient features like roads, trees, and corners, which better emphasizes co-visible regions of cross-view images. 

In addition, we selected several scenes to display the top-4 retrieval results based on feature similarity, as shown in Figure \ref{fig:visualization_retrieval}. True-matched results are highlighted with green boxes, while false-matched results are indicated with red boxes. The figure illustrates that our proposed AuxGeo effectively establishes correspondences between cross-view images. However, it may encounter challenges in geo-localization for regions lacking distinctive features.

\section{Conclusion}
\label{Conclusion}
In this work, we introduce the concept of decentrality and propose DReSS, a novel dataset for cross-view geo-localization that allows for a thorough investigation of decentrality issues. This dataset comprises cross-view images from eight global cities, representing diverse styles and encompassing extensive regions, including urban, suburban, and rural areas. To tackle both the cross-view problem and the issue of decentrality, we propose a novel method called AuxGeo. AuxGeo employs a multi-metric optimization strategy incorporating two novel auxiliary modules designed to enhance the backbone's representational ability. Extensive experiments demonstrate that AuxGeo achieves state-of-the-art performance on our proposed DReSS as well as on public datasets such as CVUSA, CVACT, and VIGOR. Importantly, AuxGeo incurs no additional pre-processes or computational costs during inference. While AuxGeo achieves a favorable balance between localization accuracy and efficiency, it falls short in addressing the low accuracy of CVGL under conditions of extremely high decentrality. This limitation highlights an important avenue for future research.

\section*{Declaration of Competing Interest}
The authors declare that they have no known competing financial interests or personal relationships that could have appeared to influence the work reported in this paper.

\section*{Acknowledgment}
This work was supported by the National Natural Science Foundation of China (Grants 42471470,  42030102, and 42192583), Ant Group and the Supercomputing Center of Wuhan University.

% \appendix
% \input{appendix.tex}

\printcredits

\bibliographystyle{cas-model2-names}
% \bibliographystyle{elsarticle-harv}
% \bibliographystyle{elsarticle-num-names}

\bibliography{reference}

\begin{thebibliography}{47}
\expandafter\ifx\csname natexlab\endcsname\relax\def\natexlab#1{#1}\fi
\providecommand{\url}[1]{\texttt{#1}}
\providecommand{\href}[2]{#2}
\providecommand{\path}[1]{#1}
\providecommand{\DOIprefix}{doi:}
\providecommand{\ArXivprefix}{arXiv:}
\providecommand{\URLprefix}{URL: }
\providecommand{\Pubmedprefix}{pmid:}
\providecommand{\doi}[1]{\href{http://dx.doi.org/#1}{\path{#1}}}
\providecommand{\Pubmed}[1]{\href{pmid:#1}{\path{#1}}}
\providecommand{\bibinfo}[2]{#2}
\ifx\xfnm\relax \def\xfnm[#1]{\unskip,\space#1}\fi
%Type = Article
\bibitem[{Cheng et~al.(2018)Cheng, Yuan, Xia, Chen, Chen, Yang, Ma and Li}]{cheng2018crowd}
\bibinfo{author}{Cheng, L.}, \bibinfo{author}{Yuan, Y.}, \bibinfo{author}{Xia, N.}, \bibinfo{author}{Chen, S.}, \bibinfo{author}{Chen, Y.}, \bibinfo{author}{Yang, K.}, \bibinfo{author}{Ma, L.}, \bibinfo{author}{Li, M.}, \bibinfo{year}{2018}.
\newblock \bibinfo{title}{Crowd-sourced pictures geo-localization method based on street view images and 3d reconstruction}.
\newblock \bibinfo{journal}{ISPRS journal of photogrammetry and remote sensing} \bibinfo{volume}{141}, \bibinfo{pages}{72--85}.
%Type = Inproceedings
\bibitem[{Chiu et~al.(2018)Chiu, Murali, Villamil, Kessler, Samarasekera and Kumar}]{chiu2018augmented}
\bibinfo{author}{Chiu, H.P.}, \bibinfo{author}{Murali, V.}, \bibinfo{author}{Villamil, R.}, \bibinfo{author}{Kessler, G.D.}, \bibinfo{author}{Samarasekera, S.}, \bibinfo{author}{Kumar, R.}, \bibinfo{year}{2018}.
\newblock \bibinfo{title}{Augmented reality driving using semantic geo-registration}, in: \bibinfo{booktitle}{2018 IEEE Conference on Virtual Reality and 3D User Interfaces (VR)}, \bibinfo{organization}{IEEE}. pp. \bibinfo{pages}{423--430}.
%Type = Inproceedings
\bibitem[{Deuser et~al.(2023)Deuser, Habel and Oswald}]{deuser2023sample4geo}
\bibinfo{author}{Deuser, F.}, \bibinfo{author}{Habel, K.}, \bibinfo{author}{Oswald, N.}, \bibinfo{year}{2023}.
\newblock \bibinfo{title}{Sample4geo: Hard negative sampling for cross-view geo-localisation}, in: \bibinfo{booktitle}{Proceedings of the IEEE/CVF International Conference on Computer Vision}, pp. \bibinfo{pages}{16847--16856}.
%Type = Article
\bibitem[{Elhashash and Qin(2022)}]{elhashash2022cross}
\bibinfo{author}{Elhashash, M.}, \bibinfo{author}{Qin, R.}, \bibinfo{year}{2022}.
\newblock \bibinfo{title}{Cross-view slam solver: Global pose estimation of monocular ground-level video frames for 3d reconstruction using a reference 3d model from satellite images}.
\newblock \bibinfo{journal}{ISPRS Journal of Photogrammetry and Remote Sensing} \bibinfo{volume}{188}, \bibinfo{pages}{62--74}.
%Type = Misc
\bibitem[{{Esri}(2024)}]{aerialview}
\bibinfo{author}{{Esri}}, \bibinfo{year}{2024}.
\newblock \bibinfo{title}{Esri world imagery}.
\newblock \bibinfo{howpublished}{\url{https://www.arcgis.com/home/item.html?id=10df2279f9684e4a9f6a7f08febac2a9}}.
%Type = Article
\bibitem[{H{\"a}ne et~al.(2017)H{\"a}ne, Heng, Lee, Fraundorfer, Furgale, Sattler and Pollefeys}]{hane20173d}
\bibinfo{author}{H{\"a}ne, C.}, \bibinfo{author}{Heng, L.}, \bibinfo{author}{Lee, G.H.}, \bibinfo{author}{Fraundorfer, F.}, \bibinfo{author}{Furgale, P.}, \bibinfo{author}{Sattler, T.}, \bibinfo{author}{Pollefeys, M.}, \bibinfo{year}{2017}.
\newblock \bibinfo{title}{3d visual perception for self-driving cars using a multi-camera system: Calibration, mapping, localization, and obstacle detection}.
\newblock \bibinfo{journal}{Image and Vision Computing} \bibinfo{volume}{68}, \bibinfo{pages}{14--27}.
%Type = Article
\bibitem[{He et~al.(2024)He, Ye, Tang, Liu and Chen}]{he2024advlut}
\bibinfo{author}{He, Y.}, \bibinfo{author}{Ye, D.}, \bibinfo{author}{Tang, L.}, \bibinfo{author}{Liu, Z.}, \bibinfo{author}{Chen, C.}, \bibinfo{year}{2024}.
\newblock \bibinfo{title}{Advlut: Cloaking geographic location with semantic-based adversarial 3d lookup tables}.
\newblock \bibinfo{journal}{IEEE Internet of Things Journal} .
%Type = Inproceedings
\bibitem[{Hu et~al.(2018)Hu, Feng, Nguyen and Lee}]{hu2018cvm}
\bibinfo{author}{Hu, S.}, \bibinfo{author}{Feng, M.}, \bibinfo{author}{Nguyen, R.M.}, \bibinfo{author}{Lee, G.H.}, \bibinfo{year}{2018}.
\newblock \bibinfo{title}{Cvm-net: Cross-view matching network for image-based ground-to-aerial geo-localization}, in: \bibinfo{booktitle}{Proceedings of the IEEE Conference on Computer Vision and Pattern Recognition}, pp. \bibinfo{pages}{7258--7267}.
%Type = Article
\bibitem[{Huang et~al.(2024)Huang, Zhou, Zhao and Gan}]{huang2024cv}
\bibinfo{author}{Huang, G.}, \bibinfo{author}{Zhou, Y.}, \bibinfo{author}{Zhao, L.}, \bibinfo{author}{Gan, W.}, \bibinfo{year}{2024}.
\newblock \bibinfo{title}{Cv-cities: Advancing cross-view geo-localization in global cities}.
\newblock \bibinfo{journal}{IEEE Journal of Selected Topics in Applied Earth Observations and Remote Sensing} .
%Type = Inproceedings
\bibitem[{Kim and Walter(2017)}]{kim2017satellite}
\bibinfo{author}{Kim, D.K.}, \bibinfo{author}{Walter, M.R.}, \bibinfo{year}{2017}.
\newblock \bibinfo{title}{Satellite image-based localization via learned embeddings}, in: \bibinfo{booktitle}{2017 IEEE international conference on robotics and automation (ICRA)}, \bibinfo{organization}{IEEE}. pp. \bibinfo{pages}{2073--2080}.
%Type = Inproceedings
\bibitem[{Li et~al.(2024a)Li, Qian and Xia}]{li2024unleashing}
\bibinfo{author}{Li, G.}, \bibinfo{author}{Qian, M.}, \bibinfo{author}{Xia, G.S.}, \bibinfo{year}{2024}a.
\newblock \bibinfo{title}{Unleashing unlabeled data: A paradigm for cross-view geo-localization}, in: \bibinfo{booktitle}{Proceedings of the IEEE/CVF Conference on Computer Vision and Pattern Recognition}, pp. \bibinfo{pages}{16719--16729}.
%Type = Article
\bibitem[{Li et~al.(2024b)Li, Deuser, Yina, Luo, Walther, Mai, Huang and Werner}]{li2024cross}
\bibinfo{author}{Li, H.}, \bibinfo{author}{Deuser, F.}, \bibinfo{author}{Yina, W.}, \bibinfo{author}{Luo, X.}, \bibinfo{author}{Walther, P.}, \bibinfo{author}{Mai, G.}, \bibinfo{author}{Huang, W.}, \bibinfo{author}{Werner, M.}, \bibinfo{year}{2024}b.
\newblock \bibinfo{title}{Cross-view geolocalization and disaster mapping with street-view and vhr satellite imagery: A case study of hurricane ian}.
\newblock \bibinfo{journal}{arXiv preprint arXiv:2408.06761} .
%Type = Misc
\bibitem[{Li et~al.(2024c)Li, Xu, Yang, Yu and Xia}]{li2024learning}
\bibinfo{author}{Li, H.}, \bibinfo{author}{Xu, C.}, \bibinfo{author}{Yang, W.}, \bibinfo{author}{Yu, H.}, \bibinfo{author}{Xia, G.S.}, \bibinfo{year}{2024}c.
\newblock \bibinfo{title}{Learning cross-view visual geo-localization without ground truth}.
\newblock \href{http://arxiv.org/abs/2403.12702}{\tt arXiv:2403.12702}.
%Type = Inproceedings
\bibitem[{Li et~al.(2023)Li, Lai, Xu, Xiangli, Yu, He, Xia and Lin}]{li2023omnicity}
\bibinfo{author}{Li, W.}, \bibinfo{author}{Lai, Y.}, \bibinfo{author}{Xu, L.}, \bibinfo{author}{Xiangli, Y.}, \bibinfo{author}{Yu, J.}, \bibinfo{author}{He, C.}, \bibinfo{author}{Xia, G.S.}, \bibinfo{author}{Lin, D.}, \bibinfo{year}{2023}.
\newblock \bibinfo{title}{Omnicity: Omnipotent city understanding with multi-level and multi-view images}, in: \bibinfo{booktitle}{Proceedings of the IEEE/CVF Conference on Computer Vision and Pattern Recognition}, pp. \bibinfo{pages}{17397--17407}.
%Type = Inproceedings
\bibitem[{Lin et~al.(2013)Lin, Belongie and Hays}]{lin2013cross}
\bibinfo{author}{Lin, T.Y.}, \bibinfo{author}{Belongie, S.}, \bibinfo{author}{Hays, J.}, \bibinfo{year}{2013}.
\newblock \bibinfo{title}{Cross-view image geolocalization}, in: \bibinfo{booktitle}{Proceedings of the IEEE Conference on Computer Vision and Pattern Recognition}, pp. \bibinfo{pages}{891--898}.
%Type = Article
\bibitem[{Ling and Qin(2022)}]{ling2022graph}
\bibinfo{author}{Ling, X.}, \bibinfo{author}{Qin, R.}, \bibinfo{year}{2022}.
\newblock \bibinfo{title}{A graph-matching approach for cross-view registration of over-view and street-view based point clouds}.
\newblock \bibinfo{journal}{ISPRS Journal of Photogrammetry and Remote Sensing} \bibinfo{volume}{185}, \bibinfo{pages}{2--15}.
%Type = Inproceedings
\bibitem[{Liu and Li(2019)}]{liu2019lending}
\bibinfo{author}{Liu, L.}, \bibinfo{author}{Li, H.}, \bibinfo{year}{2019}.
\newblock \bibinfo{title}{Lending orientation to neural networks for cross-view geo-localization}, in: \bibinfo{booktitle}{Proceedings of the IEEE/CVF conference on computer vision and pattern recognition}, pp. \bibinfo{pages}{5624--5633}.
%Type = Inproceedings
\bibitem[{Liu et~al.(2022)Liu, Mao, Wu, Feichtenhofer, Darrell and Xie}]{liu2022convnet}
\bibinfo{author}{Liu, Z.}, \bibinfo{author}{Mao, H.}, \bibinfo{author}{Wu, C.Y.}, \bibinfo{author}{Feichtenhofer, C.}, \bibinfo{author}{Darrell, T.}, \bibinfo{author}{Xie, S.}, \bibinfo{year}{2022}.
\newblock \bibinfo{title}{A convnet for the 2020s}, in: \bibinfo{booktitle}{Proceedings of the IEEE/CVF conference on computer vision and pattern recognition}, pp. \bibinfo{pages}{11976--11986}.
%Type = Inproceedings
\bibitem[{Lu et~al.(2020)Lu, Li, Cui, Oswald, Pollefeys and Qin}]{lu2020geometry}
\bibinfo{author}{Lu, X.}, \bibinfo{author}{Li, Z.}, \bibinfo{author}{Cui, Z.}, \bibinfo{author}{Oswald, M.R.}, \bibinfo{author}{Pollefeys, M.}, \bibinfo{author}{Qin, R.}, \bibinfo{year}{2020}.
\newblock \bibinfo{title}{Geometry-aware satellite-to-ground image synthesis for urban areas}, in: \bibinfo{booktitle}{Proceedings of the IEEE/CVF Conference on Computer Vision and Pattern Recognition}, pp. \bibinfo{pages}{859--867}.
%Type = Inproceedings
\bibitem[{McManus et~al.(2014)McManus, Churchill, Maddern, Stewart and Newman}]{mcmanus2014shady}
\bibinfo{author}{McManus, C.}, \bibinfo{author}{Churchill, W.}, \bibinfo{author}{Maddern, W.}, \bibinfo{author}{Stewart, A.D.}, \bibinfo{author}{Newman, P.}, \bibinfo{year}{2014}.
\newblock \bibinfo{title}{Shady dealings: Robust, long-term visual localisation using illumination invariance}, in: \bibinfo{booktitle}{2014 IEEE international conference on robotics and automation (ICRA)}, \bibinfo{organization}{IEEE}. pp. \bibinfo{pages}{901--906}.
%Type = Inproceedings
\bibitem[{Regmi and Borji(2018)}]{regmi2018cross}
\bibinfo{author}{Regmi, K.}, \bibinfo{author}{Borji, A.}, \bibinfo{year}{2018}.
\newblock \bibinfo{title}{Cross-view image synthesis using conditional gans}, in: \bibinfo{booktitle}{Proceedings of the IEEE conference on Computer Vision and Pattern Recognition}, pp. \bibinfo{pages}{3501--3510}.
%Type = Inproceedings
\bibitem[{Regmi and Shah(2019)}]{regmi2019bridging}
\bibinfo{author}{Regmi, K.}, \bibinfo{author}{Shah, M.}, \bibinfo{year}{2019}.
\newblock \bibinfo{title}{Bridging the domain gap for ground-to-aerial image matching}, in: \bibinfo{booktitle}{Proceedings of the IEEE/CVF International Conference on Computer Vision}, pp. \bibinfo{pages}{470--479}.
%Type = Article
\bibitem[{Shi et~al.(2022)Shi, Campbell, Yu and Li}]{shi2022geometry}
\bibinfo{author}{Shi, Y.}, \bibinfo{author}{Campbell, D.}, \bibinfo{author}{Yu, X.}, \bibinfo{author}{Li, H.}, \bibinfo{year}{2022}.
\newblock \bibinfo{title}{Geometry-guided street-view panorama synthesis from satellite imagery}.
\newblock \bibinfo{journal}{IEEE Transactions on Pattern Analysis and Machine Intelligence} \bibinfo{volume}{44}, \bibinfo{pages}{10009--10022}.
%Type = Article
\bibitem[{Shi et~al.(2019)Shi, Liu, Yu and Li}]{shi2019spatial}
\bibinfo{author}{Shi, Y.}, \bibinfo{author}{Liu, L.}, \bibinfo{author}{Yu, X.}, \bibinfo{author}{Li, H.}, \bibinfo{year}{2019}.
\newblock \bibinfo{title}{Spatial-aware feature aggregation for image based cross-view geo-localization}.
\newblock \bibinfo{journal}{Advances in Neural Information Processing Systems} \bibinfo{volume}{32}.
%Type = Inproceedings
\bibitem[{Shi et~al.(2020a)Shi, Yu, Campbell and Li}]{shi2020looking}
\bibinfo{author}{Shi, Y.}, \bibinfo{author}{Yu, X.}, \bibinfo{author}{Campbell, D.}, \bibinfo{author}{Li, H.}, \bibinfo{year}{2020}a.
\newblock \bibinfo{title}{Where am i looking at? joint location and orientation estimation by cross-view matching}, in: \bibinfo{booktitle}{Proceedings of the IEEE/CVF Conference on Computer Vision and Pattern Recognition}, pp. \bibinfo{pages}{4064--4072}.
%Type = Inproceedings
\bibitem[{Shi et~al.(2020b)Shi, Yu, Liu, Zhang and Li}]{shi2020optimal}
\bibinfo{author}{Shi, Y.}, \bibinfo{author}{Yu, X.}, \bibinfo{author}{Liu, L.}, \bibinfo{author}{Zhang, T.}, \bibinfo{author}{Li, H.}, \bibinfo{year}{2020}b.
\newblock \bibinfo{title}{Optimal feature transport for cross-view image geo-localization}, in: \bibinfo{booktitle}{Proceedings of the AAAI Conference on Artificial Intelligence}, pp. \bibinfo{pages}{11990--11997}.
%Type = Inproceedings
\bibitem[{Sun et~al.(2019)Sun, Chen, Zhu and Jiang}]{sun2019geocapsnet}
\bibinfo{author}{Sun, B.}, \bibinfo{author}{Chen, C.}, \bibinfo{author}{Zhu, Y.}, \bibinfo{author}{Jiang, J.}, \bibinfo{year}{2019}.
\newblock \bibinfo{title}{Geocapsnet: Ground to aerial view image geo-localization using capsule network}, in: \bibinfo{booktitle}{2019 IEEE International Conference on Multimedia and Expo (ICME)}, \bibinfo{organization}{IEEE}. pp. \bibinfo{pages}{742--747}.
%Type = Inproceedings
\bibitem[{Tian et~al.(2017)Tian, Chen and Shah}]{tian2017cross}
\bibinfo{author}{Tian, Y.}, \bibinfo{author}{Chen, C.}, \bibinfo{author}{Shah, M.}, \bibinfo{year}{2017}.
\newblock \bibinfo{title}{Cross-view image matching for geo-localization in urban environments}, in: \bibinfo{booktitle}{Proceedings of the IEEE Conference on Computer Vision and Pattern Recognition}, pp. \bibinfo{pages}{3608--3616}.
%Type = Inproceedings
\bibitem[{Toker et~al.(2021)Toker, Zhou, Maximov and Leal-Taix{\'e}}]{toker2021coming}
\bibinfo{author}{Toker, A.}, \bibinfo{author}{Zhou, Q.}, \bibinfo{author}{Maximov, M.}, \bibinfo{author}{Leal-Taix{\'e}, L.}, \bibinfo{year}{2021}.
\newblock \bibinfo{title}{Coming down to earth: Satellite-to-street view synthesis for geo-localization}, in: \bibinfo{booktitle}{Proceedings of the IEEE/CVF Conference on Computer Vision and Pattern Recognition}, pp. \bibinfo{pages}{6488--6497}.
%Type = Inproceedings
\bibitem[{Vo and Hays(2016)}]{vo2016localizing}
\bibinfo{author}{Vo, N.N.}, \bibinfo{author}{Hays, J.}, \bibinfo{year}{2016}.
\newblock \bibinfo{title}{Localizing and orienting street views using overhead imagery}, in: \bibinfo{booktitle}{Computer Vision--ECCV 2016: 14th European Conference, Amsterdam, The Netherlands, October 11--14, 2016, Proceedings, Part I 14}, \bibinfo{organization}{Springer}. pp. \bibinfo{pages}{494--509}.
%Type = Article
\bibitem[{Wan et~al.(2016)Wan, Liu, Yan and Morgan}]{wan2016illumination}
\bibinfo{author}{Wan, X.}, \bibinfo{author}{Liu, J.}, \bibinfo{author}{Yan, H.}, \bibinfo{author}{Morgan, G.L.}, \bibinfo{year}{2016}.
\newblock \bibinfo{title}{Illumination-invariant image matching for autonomous uav localisation based on optical sensing}.
\newblock \bibinfo{journal}{ISPRS Journal of Photogrammetry and Remote Sensing} \bibinfo{volume}{119}, \bibinfo{pages}{198--213}.
%Type = Article
\bibitem[{Wang et~al.(2023)Wang, Li and Sun}]{wang2023dehi}
\bibinfo{author}{Wang, T.}, \bibinfo{author}{Li, J.}, \bibinfo{author}{Sun, C.}, \bibinfo{year}{2023}.
\newblock \bibinfo{title}{Dehi: A decoupled hierarchical architecture for unaligned ground-to-aerial geo-localization}.
\newblock \bibinfo{journal}{IEEE Transactions on Circuits and Systems for Video Technology} .
%Type = Article
\bibitem[{Wang et~al.(2024)Wang, Xu, Cui, Wan and Zhang}]{wang2024fine}
\bibinfo{author}{Wang, X.}, \bibinfo{author}{Xu, R.}, \bibinfo{author}{Cui, Z.}, \bibinfo{author}{Wan, Z.}, \bibinfo{author}{Zhang, Y.}, \bibinfo{year}{2024}.
\newblock \bibinfo{title}{Fine-grained cross-view geo-localization using a correlation-aware homography estimator}.
\newblock \bibinfo{journal}{Advances in Neural Information Processing Systems} \bibinfo{volume}{36}.
%Type = Inproceedings
\bibitem[{Workman et~al.(2015)Workman, Souvenir and Jacobs}]{workman2015wide}
\bibinfo{author}{Workman, S.}, \bibinfo{author}{Souvenir, R.}, \bibinfo{author}{Jacobs, N.}, \bibinfo{year}{2015}.
\newblock \bibinfo{title}{Wide-area image geolocalization with aerial reference imagery}, in: \bibinfo{booktitle}{Proceedings of the IEEE International Conference on Computer Vision}, pp. \bibinfo{pages}{3961--3969}.
%Type = Article
\bibitem[{Wu et~al.(2024)Wu, Wan, Zheng, Zhang, Wang and Zhao}]{wu2024camp}
\bibinfo{author}{Wu, Q.}, \bibinfo{author}{Wan, Y.}, \bibinfo{author}{Zheng, Z.}, \bibinfo{author}{Zhang, Y.}, \bibinfo{author}{Wang, G.}, \bibinfo{author}{Zhao, Z.}, \bibinfo{year}{2024}.
\newblock \bibinfo{title}{Camp: A cross-view geo-localization method using contrastive attributes mining and position-aware partitioning}.
\newblock \bibinfo{journal}{IEEE Transactions on Geoscience and Remote Sensing} .
%Type = Article
\bibitem[{Xia et~al.(2024)Xia, Wan, Zheng, Zhang and Deng}]{xia2024enhancing}
\bibinfo{author}{Xia, P.}, \bibinfo{author}{Wan, Y.}, \bibinfo{author}{Zheng, Z.}, \bibinfo{author}{Zhang, Y.}, \bibinfo{author}{Deng, J.}, \bibinfo{year}{2024}.
\newblock \bibinfo{title}{Enhancing cross-view geo-localization with domain alignment and scene consistency}.
\newblock \bibinfo{journal}{IEEE Transactions on Circuits and Systems for Video Technology} .
%Type = Article
\bibitem[{Yang et~al.(2021)Yang, Lu and Zhu}]{yang2021cross}
\bibinfo{author}{Yang, H.}, \bibinfo{author}{Lu, X.}, \bibinfo{author}{Zhu, Y.}, \bibinfo{year}{2021}.
\newblock \bibinfo{title}{Cross-view geo-localization with layer-to-layer transformer}.
\newblock \bibinfo{journal}{Advances in Neural Information Processing Systems} \bibinfo{volume}{34}, \bibinfo{pages}{29009--29020}.
%Type = Inproceedings
\bibitem[{Ye et~al.(2024a)Ye, Luo, Yu, Zhong, Zheng, He and Li}]{ye2024sg}
\bibinfo{author}{Ye, J.}, \bibinfo{author}{Luo, Q.}, \bibinfo{author}{Yu, J.}, \bibinfo{author}{Zhong, H.}, \bibinfo{author}{Zheng, Z.}, \bibinfo{author}{He, C.}, \bibinfo{author}{Li, W.}, \bibinfo{year}{2024}a.
\newblock \bibinfo{title}{Sg-bev: Satellite-guided bev fusion for cross-view semantic segmentation}, in: \bibinfo{booktitle}{Proceedings of the IEEE/CVF Conference on Computer Vision and Pattern Recognition}, pp. \bibinfo{pages}{27748--27757}.
%Type = Inproceedings
\bibitem[{Ye et~al.(2025)Ye, Lv, Li, Yu, Yang, Zhong and He}]{ye2025cross}
\bibinfo{author}{Ye, J.}, \bibinfo{author}{Lv, Z.}, \bibinfo{author}{Li, W.}, \bibinfo{author}{Yu, J.}, \bibinfo{author}{Yang, H.}, \bibinfo{author}{Zhong, H.}, \bibinfo{author}{He, C.}, \bibinfo{year}{2025}.
\newblock \bibinfo{title}{Cross-view image geo-localization with panorama-bev co-retrieval network}, in: \bibinfo{booktitle}{European Conference on Computer Vision}, \bibinfo{organization}{Springer}. pp. \bibinfo{pages}{74--90}.
%Type = Article
\bibitem[{Ye et~al.(2024b)Ye, Luo and Lin}]{ye2024coarse}
\bibinfo{author}{Ye, Q.}, \bibinfo{author}{Luo, J.}, \bibinfo{author}{Lin, Y.}, \bibinfo{year}{2024}b.
\newblock \bibinfo{title}{A coarse-to-fine visual geo-localization method for gnss-denied uav with oblique-view imagery}.
\newblock \bibinfo{journal}{ISPRS Journal of Photogrammetry and Remote Sensing} \bibinfo{volume}{212}, \bibinfo{pages}{306--322}.
%Type = Inproceedings
\bibitem[{Zhai et~al.(2017)Zhai, Bessinger, Workman and Jacobs}]{zhai2017predicting}
\bibinfo{author}{Zhai, M.}, \bibinfo{author}{Bessinger, Z.}, \bibinfo{author}{Workman, S.}, \bibinfo{author}{Jacobs, N.}, \bibinfo{year}{2017}.
\newblock \bibinfo{title}{Predicting ground-level scene layout from aerial imagery}, in: \bibinfo{booktitle}{Proceedings of the IEEE Conference on Computer Vision and Pattern Recognition}, pp. \bibinfo{pages}{867--875}.
%Type = Inproceedings
\bibitem[{Zhang and Zhu(2024)}]{zhang2024aligning}
\bibinfo{author}{Zhang, Q.}, \bibinfo{author}{Zhu, Y.}, \bibinfo{year}{2024}.
\newblock \bibinfo{title}{Aligning geometric spatial layout in cross-view geo-localization via feature recombination}, in: \bibinfo{booktitle}{Proceedings of the AAAI Conference on Artificial Intelligence}, pp. \bibinfo{pages}{7251--7259}.
%Type = Inproceedings
\bibitem[{Zhang et~al.(2023)Zhang, Li, Sultani, Zhou and Wshah}]{zhang2023cross}
\bibinfo{author}{Zhang, X.}, \bibinfo{author}{Li, X.}, \bibinfo{author}{Sultani, W.}, \bibinfo{author}{Zhou, Y.}, \bibinfo{author}{Wshah, S.}, \bibinfo{year}{2023}.
\newblock \bibinfo{title}{Cross-view geo-localization via learning disentangled geometric layout correspondence}, in: \bibinfo{booktitle}{Proceedings of the AAAI Conference on Artificial Intelligence}, pp. \bibinfo{pages}{3480--3488}.
%Type = Inproceedings
\bibitem[{Zheng et~al.(2020)Zheng, Wei and Yang}]{zheng2020university}
\bibinfo{author}{Zheng, Z.}, \bibinfo{author}{Wei, Y.}, \bibinfo{author}{Yang, Y.}, \bibinfo{year}{2020}.
\newblock \bibinfo{title}{University-1652: A multi-view multi-source benchmark for drone-based geo-localization}, in: \bibinfo{booktitle}{Proceedings of the 28th ACM international conference on Multimedia}, pp. \bibinfo{pages}{1395--1403}.
%Type = Inproceedings
\bibitem[{Zhu et~al.(2022)Zhu, Shah and Chen}]{zhu2022transgeo}
\bibinfo{author}{Zhu, S.}, \bibinfo{author}{Shah, M.}, \bibinfo{author}{Chen, C.}, \bibinfo{year}{2022}.
\newblock \bibinfo{title}{Transgeo: Transformer is all you need for cross-view image geo-localization}, in: \bibinfo{booktitle}{Proceedings of the IEEE/CVF Conference on Computer Vision and Pattern Recognition}, pp. \bibinfo{pages}{1162--1171}.
%Type = Inproceedings
\bibitem[{Zhu et~al.(2021)Zhu, Yang and Chen}]{zhu2021vigor}
\bibinfo{author}{Zhu, S.}, \bibinfo{author}{Yang, T.}, \bibinfo{author}{Chen, C.}, \bibinfo{year}{2021}.
\newblock \bibinfo{title}{Vigor: Cross-view image geo-localization beyond one-to-one retrieval}, in: \bibinfo{booktitle}{Proceedings of the IEEE/CVF Conference on Computer Vision and Pattern Recognition}, pp. \bibinfo{pages}{3640--3649}.
%Type = Misc
\bibitem[{Zhu et~al.(2023)Zhu, Yang, Lu and Huang}]{zhu2023simple}
\bibinfo{author}{Zhu, Y.}, \bibinfo{author}{Yang, H.}, \bibinfo{author}{Lu, Y.}, \bibinfo{author}{Huang, Q.}, \bibinfo{year}{2023}.
\newblock \bibinfo{title}{Simple, effective and general: A new backbone for cross-view image geo-localization}.
\newblock \href{http://arxiv.org/abs/2302.01572}{\tt arXiv:2302.01572}.

\end{thebibliography}


% \input{bio.tex}

\end{document}